\documentclass{article}

\usepackage[dblblindworkshop, final]{neurips_2026}
\setcitestyle{numbers,square,comma}

\usepackage[utf8]{inputenc} 
\usepackage[T1]{fontenc}    
\usepackage[dvipsnames]{xcolor}
\usepackage[linkcolor=BrickRed,colorlinks,citecolor=NavyBlue,urlcolor=NavyBlue,pagebackref=true]{hyperref}
\usepackage{url}            
\usepackage{booktabs}       
\usepackage{amsfonts}       
\usepackage{nicefrac}       
\usepackage{microtype}      
\usepackage{multirow, makecell}
\usepackage{amsmath}
\usepackage{graphicx}
\usepackage{subcaption}
\usepackage{cleveref}

\newcommand{\eg}{\textit{e.g.}~}
\newcommand{\ie}{\textit{i.e.}~}

\newcommand{\pair}{\ensuremath{\times}}

\title{Towards Consistent Geospatial Representations: Aligning One Encoder to Many Foundation Models}
\workshoptitle{2nd Workshop on Advances in Representation Learning for Earth Observation}

\author{%
  Arthur Ouaknine \\
  McGill University\\
  Mila -- Quebec AI Institute\\
  \texttt{arthur.ouaknine@mila.quebec} \\
  \And
  David Rolnick \\
  McGill University\\
  Mila -- Quebec AI Institute\\
}

\begin{document}

\maketitle

\begin{abstract}
Geospatial foundation models (GFMs) differ widely in pretraining objective, modality and resolution, and none leads across benchmarks. 
We ask whether their representations are complementary, and whether a single encoder can inherit that complementarity. 
We introduce \textsc{Pangaea-Latents}, an extension of the \textsc{Pangaea} benchmark providing frozen embeddings of $5$ GFMs over its $11$ datasets, along with reproduced baselines, and find GFM embeddings to be neither redundant nor disjoint. 
We then propose \textsc{GeoMixture}, a preliminary framework to align a ViT to several GFMs per dataset through a multi-representation alignment (mREPA) objective. 
Our encoder ties the best GFM in average rank across nine \textsc{Pangaea} datasets, trading peak performance for cross-task consistency.
\end{abstract}

\section{Introduction}

GFMs have the potential to transform Earth and environmental sciences by learning processes from measurements at planetary scale \cite{zhu_foundations_2026}. 
Trained by self-supervised learning (SSL) on terabytes of satellite recordings without ground-truth supervision, they map observations from heterogeneous sensors, wavelengths, resolutions and dates onto dense Earth embeddings, reusable for arbitrary downstream tasks \cite{klemmer_earth_2026}. 
Benchmarks probe how well these frozen representations adapt across regimes, tasks and geographies \cite{lacoste_geo-bench_2023, marsocci_pangaea_2025, bountos_fomo_2025}, \textsc{Pangaea} \cite{marsocci_pangaea_2025} being widely adopted for its diversity and its evaluation framework suited to ViT backbones. 
Yet no single GFM leads across benchmarks \cite{corley_no_2026}.
One plausible reason is that pretraining objectives, modalities and architectures differ across GFMs (Sec.~\ref{sec:related_work}), each covering a different region of the downstream task landscape.

Combining GFMs encoding different information could broaden coverage at a fraction of the cost of pretraining another model \cite{plas_better_2026}. Weight-space merging \cite{hanna_geosane_2026} requires architecture compatibility and released checkpoints, excluding embedding-only models, \eg AlphaEarth \cite{brown_alphaearth_2025}; combination in latent space has neither constraint. We explore it by first introducing \textsc{Pangaea-Latents}, an extension of \textsc{Pangaea} providing the embeddings of $5$ GFMs for each of $11$ datasets, with baselines reproduced over $3$ seeds. 
We analyze their similarity and complementarity, finding that the benefit of mixing them depends on the dataset and is not predicted by their similarity. 
We then propose \textsc{GeoMixture}, a framework aligning a ViT with several GFMs through a multi-representation alignment (mREPA) objective, aiming not for the best score on every dataset but for consistency across the benchmark.


\section{Related work}
\label{sec:related_work}

\paragraph{Diversity of GFM pretraining.} GFMs explore a wide range of SSL objectives on multimodal remote sensing data: masked image modelling \cite{xiong_neural_2025, bountos_fomo_2025, jakubik_terramind_2025, forgaard_thor_2026}, contrastive and hybrid objectives \cite{fuller_croma_2023, tseng_galileo_2025, herzog_olmoearth_2025, leonardis_omnisat_2025,
tseng_olmoearth_2026, feng_tessera_2026}, and joint-embedding predictive architectures \cite{astruc_anysat_2025}. Others target modality-specific properties such as scale \cite{reed_scale-mae_2023, houdre_ramen_2025}, temporal dynamics \cite{szwarcman_prithvi-eo-20_2026}, spectral heterogeneity \cite{hong_spectralgpt_2024} or geolocation \cite{klemmer_satclip_2025}, or structuring the latent space by modality with Mixtures of Experts (MoE) \cite{zhang_skysense_2025}.

\paragraph{Complementarity of learned representations.} Whether such diverse recipes converge to similar representations is an open question, raised for vision models in general \cite{huh_position_2024}. In the geospatial setting, \citet{plas_better_2026} propose a complementarity score estimating whether two stacked GFM embeddings should improve performance on a task, a useful prior but based on linear probes that does not yield to a deployable model.
We extend this analysis to another set of GFMs within the \textsc{Pangaea} framework.

\paragraph{Combining foundation models.} In weight space, \citet{hanna_geosane_2026} sample GFM weights from a generative model, producing a model on demand at the price of sampling variability and a per-task cost. In the latent space, multi-teacher distillation yields, on natural images, a single backbone matching or exceeding its teachers \cite{ranzinger_am-radio_2024, leonardis_unic_2025}. We follow this line with a focus on representation alignment \cite{yu_representation_2025}, aligning a single ViT to $N$ GFMs through one lightweight projection per teacher.


\section{Benchmark and method}
\label{sec:benchmark_method}

\subsection{\textsc{Pangaea-Latents}}
\label{sec:pangaea_latents}

We propose \textsc{Pangaea-Latents} (Fig.~\ref{fig:teaser_pangaea_latents}), an extension of \textsc{Pangaea} \cite{marsocci_pangaea_2025} (App.~\ref{sec:app_pangaea}) providing embeddings $e_{i,j}$ for each of its $D=11$ datasets $i$ and each of $N=5$ GFMs $j$: CROMA \cite{fuller_croma_2023}, DOFA \cite{xiong_neural_2025}, Prithvi \cite{szwarcman_prithvi-eo-20_2026}, Scale-MAE \cite{reed_scale-mae_2023} and TerraMind-L \cite{jakubik_terramind_2025}, spanning distinct pretraining objectives and input modalities. Each encoder is applied once to every split under its own preprocessing and band mapping, and the embedding produced by its last layer is stored, then standardized with min-max values per GFM and per dataset for our experiments.
Precomputing the latents decouples running the GFMs from training on top of them: a student is trained without a single forward pass through any GFM, making our setup (Sec.~\ref{sec:geomixture}) tractable on one GPU. Further details are provided in App.~\ref{sec:app_pangaea_latents}.

\subsection{Similarity and complementarity scores}
\label{sec:similarity_complementarity}

Let $e_{i,k}=X \in \mathbb{R}^{n \times d_k}$ and $e_{i,l}=Y \in \mathbb{R}^{n \times d_l}$ be two GFM embeddings for a dataset $i \in \{1,\cdots,D\}$, with $n$ matched tokens over the dataset’s tiles. We quantify their similarity with the canonical correlation analysis (CCA) and with the linear centered kernel alignment (CKA) \cite{kornblith_similarity_2019}; both are detailed in App.~\ref{sec:app_similarity_complementarity}. 
As CCA gives importance to directions corresponding to noise and linear CKA is dominated by high-variance directions \cite{davari_reliability_2023} since GFMs features have highly uneven per-dimension variances, we also report a standardized variant, computed on column-standardized features,
\begin{equation}
  \mathrm{CKA_{std}}(X, Y) =
  \frac{\lVert \tilde{Y}^{\top} \tilde{X} \rVert_F^{2}}
       {\lVert \tilde{X}^{\top} \tilde{X} \rVert_F \, \lVert \tilde{Y}^{\top} \tilde{Y} \rVert_F},
  \label{eq:cka_std}
\end{equation}
where $\tilde{X}_{:,c} = \frac{X_{:,c} - \mu_c}{\sigma_c + \varepsilon}$, with $\mu_c$ and $\sigma_c$ the mean and standard deviation of column $c$ and $\varepsilon = 10^{-6}$.

The complementarity score (Eq.~\ref{eq:complementarity}) is the normalized gain of a ridge regression on concatenated embeddings over the better individual probe \citep{plas_better_2026}. Details are provided in App.~\ref{sec:app_similarity_complementarity}.


\subsection{\textsc{GeoMixture}}
\label{sec:geomixture}

\begin{figure}[t!]
\vspace{-5mm}
    \centering
  \begin{subfigure}[b]{0.26\linewidth}
    \includegraphics[width=\linewidth]{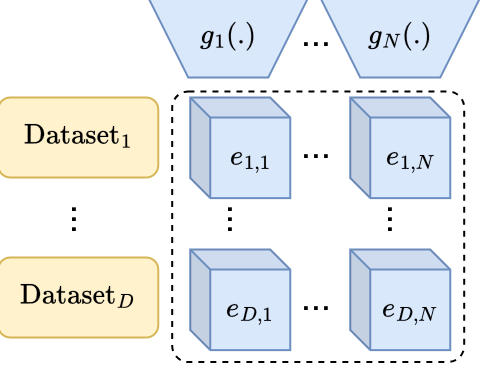}
    \caption{\textsc{Pangaea-latents}}
    \label{fig:teaser_pangaea_latents}
  \end{subfigure}
  \hspace{0.02\linewidth}
  \begin{subfigure}[b]{0.26\linewidth}
    \includegraphics[width=\linewidth]{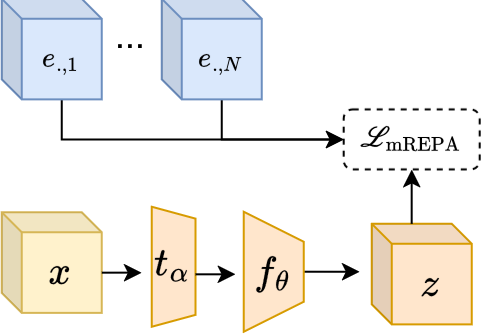}
    \caption{\textsc{GeoMixture-Naive}}
    \label{fig:geomixture_naive}
  \end{subfigure}
  \hspace{0.02\linewidth}
  \begin{subfigure}[b]{0.39\linewidth}
    \includegraphics[width=\linewidth]{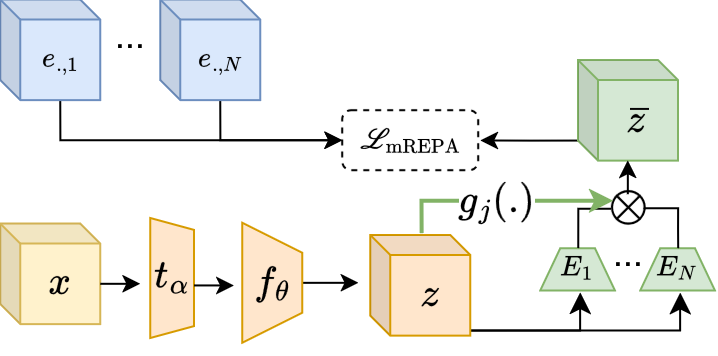}
    \caption{\textsc{GeoMixture-MoE}}
    \label{fig:geomixture_moe}
  \end{subfigure}
  \caption{\textbf{\textsc{Pangaea-Latents}} (a) (Sec.~\ref{sec:pangaea_latents}) is an extension of \textsc{Pangaea} with frozen embeddings of GFMs. \textbf{\textsc{GeoMixture}} (Sec.~\ref{sec:geomixture}) is a framework aligning a ViT to GFMs through a multi-representation alignment (mREPA) objective in a single latent (b), or in a Mixture of Experts (c). }
  \label{fig:teaset_methods}
\end{figure}

Components below are trained independently for each dataset with the same $N=5$ GFM teachers; we omit the dataset index for readability. Teachers are frozen and exploited via \textsc{Pangaea-Latents}.

\paragraph{Student encoder.} Let $x \in \mathbb{R}^{M \times w \times h}$ be the stacked input modalities, with $M$ channels and spatial size $w \times h$. A tokenizer $t_\alpha$, implemented as a single 2D convolution, maps $x$ to a token sequence encoded by a ViT-B/$16$ backbone $f_\theta$ initialized from ImageNet, producing $z \in \mathbb{R}^{d \times P}$ with $d = 768$ and $P = 196$. Let $z^{[s]} \in \mathbb{R}^{d}$ be the token at location $s$, and $e_j \in \mathbb{R}^{d_j \times P}$ the embedding of GFM $j$ for the same input, with $d_j \in \{768, 1024\}$. 
The tokenizer, backbone, experts and router (for mixture variants) and projections $p_j$ are trained by the alignment objective; teachers stay frozen.

\paragraph{Naive vs. mixture of experts.} \textsc{GeoMixture-Naive} aligns $z$ directly (Fig.~\ref{fig:geomixture_naive}). \textsc{GeoMixture-MoE} instead routes it through $N$ experts (Fig.~\ref{fig:geomixture_moe}), one per teacher, each $E_j : \mathbb{R}^{d} \rightarrow \mathbb{R}^{d}$ a $\text{1DConv} \rightarrow \text{GELU} \rightarrow \text{1DConv}$ block applied token-wise. A router $W_g \in \mathbb{R}^{N \times d}$, implemented as a $2$D convolution over the channel dimension, produces per-location weights over the experts,
\begin{equation}
  g_j(z^{[s]}) =
  \frac{m_j \exp\!\big( (W_g\, z^{[s]})_j \big)}
       {\sum_{k=1}^{N} m_k \exp\!\big( (W_g\, z^{[s]})_k \big)},
  \qquad
  \bar{z}^{[s]} = \sum_{j=1}^{N} g_j(z^{[s]})\, E_j(z^{[s]}),
  \label{eq:moe}
\end{equation}
where $m \in \{0,1\}^{N}$ masks teachers unavailable for the current sample, so weights sum to one at every location. 
Routing is input-dependent and spatially varying: different regions of a modality may draw on different teachers, while the output retains the backbone width $d$. We also report \textsc{GeoMixture-Sum}, replacing the router by $g_j = 1$, summing available experts without reweighting.

\paragraph{Multi-representation alignment.} Let $u = z$ for \textsc{Naive}, $u = \bar{z}$ for \textsc{Sum} and \textsc{MoE} versions of \textsc{GeoMixture}. Each teacher is matched through a projection $p_j : \mathbb{R}^{d} \rightarrow \mathbb{R}^{d_j}$, a single token-wise $1$D convolution that adjusts the width without adding capacity beyond the experts. We minimize
\begin{equation}
  \mathcal{L}_{\mathrm{mREPA}}\big(\{e_j\}_{j=1}^{N},\, u\big)
  = \frac{1}{P \sum_{j} m_j}
    \sum_{j=1}^{N} m_j \sum_{s=1}^{P}
    \ell\!\left(e_j^{[s]},\, p_j\!\left(u^{[s]}\right)\right),
  \label{eq:mrepa}
\end{equation}
where $\ell$ is a discrepancy between two token embeddings. 
We consider the MSE $\ell_{\mathrm{mse}}(a,b) = \lVert a - b \rVert_2^2$, the negative cosine similarity $\ell_{\cos}(a,b) = - a^\top b / (\lVert a \rVert \lVert b \rVert)$, and their sum $\ell_{\mathrm{mse}} + \ell_{\cos}$ (Tab.~\ref{tab:latent_moe_ablation}).
Eq.~\ref{eq:mrepa} costs $N$ projections rather than $N$ GFM forward passes,
and is normalized across the teachers.

\paragraph{From alignment to evaluation.} Alignment supervises only the final
block of $f_\theta$. After training, the projections $p_j$ are discarded and the
tokenizer, the backbone and, for the mixture variants, the experts and router
are evaluated frozen under the \textsc{Pangaea} protocol
(App.~\ref{app:pangaea_protocol}), with a UPerNet decoder reading intermediate
stages of $f_\theta$ together with $z$ or $\bar{z}$. Intermediate
representations are thus shaped only indirectly, and the mixture belongs to the
evaluated encoder rather than to training alone.


\section{Experiments and results}
\label{sec:results}

\begin{figure}[t]
\vspace{-5mm}
    \centering
      \makebox[\linewidth][c]{%
  \resizebox{1.2\linewidth}{!}{

  \begin{subfigure}[b]{0.245\linewidth}
    \includegraphics[trim=12 38 11 46, clip, width=\linewidth]{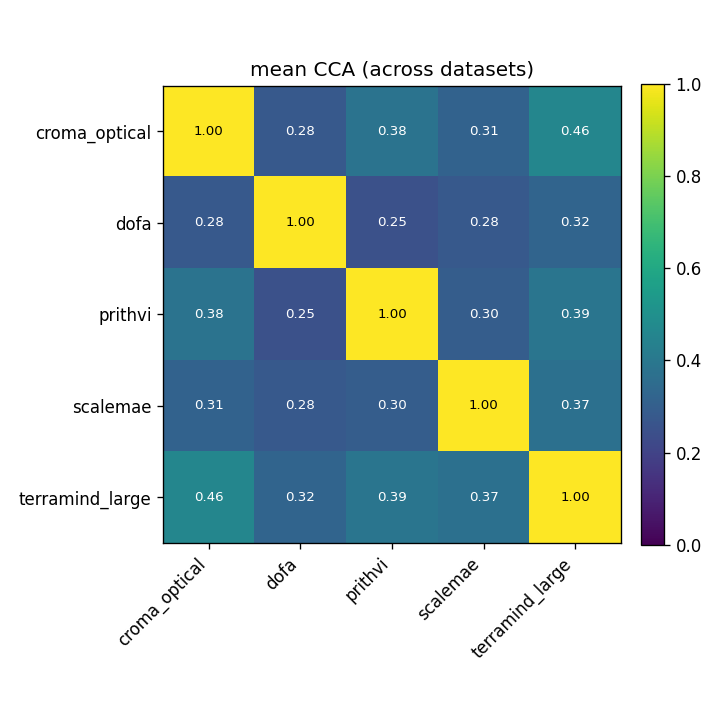}
    \caption{CCA}
    \label{fig:summary_scores_a}
  \end{subfigure}
  \begin{subfigure}[b]{0.245\linewidth}
    \includegraphics[trim=12 38 11 46, clip, width=\linewidth]{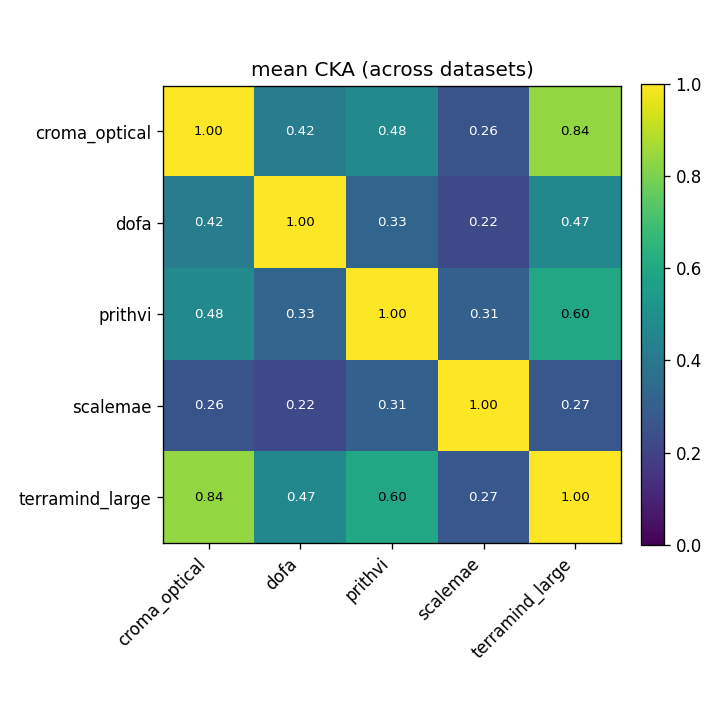}
    \caption{CKA}
    \label{fig:summary_scores_b}
  \end{subfigure}
  \begin{subfigure}[b]{0.245\linewidth}
    \includegraphics[trim=12 38 11 48, clip, width=\linewidth]{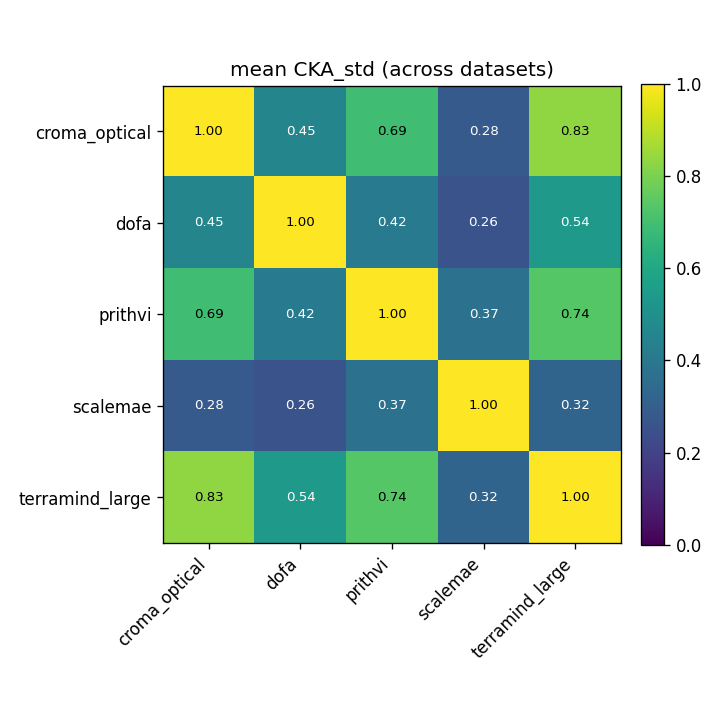}
    \caption{Standardized CKA}
    \label{fig:summary_scores_c}
  \end{subfigure}
  \begin{subfigure}[b]{0.245\linewidth}
    \includegraphics[trim=12 38 11 52, clip, width=\linewidth]{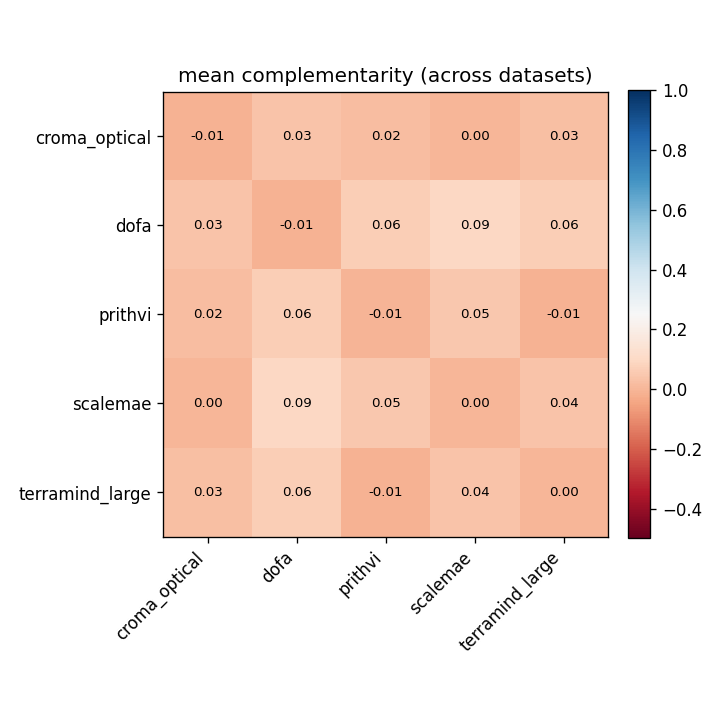}
    \caption{Complementarity}
    \label{fig:summary_scores_d}
  \end{subfigure}
  }}
  \caption{Pairwise similarity (Eqs.~\ref{eq:cca},~\ref{eq:cka},~\ref{eq:cka_std}) and
  complementarity (Eq.~\ref{eq:complementarity}) between the $5$ GFM embeddings, averaged
  over the $11$ datasets of \textsc{Pangaea-Latents}; per-dataset scores in
  App.~\ref{sec:app_scores}.}
  \label{fig:summary_scores}
\end{figure}

\paragraph{Similarity and complementarity.}
Fig.~\ref{fig:summary_scores} reports scores aggregated on 11 \textsc{Pangaea-Latents} datasets (per-dataset results in App.~\ref{sec:app_scores}), using standardized CKA as similarity given the limitations of CCA and CKA (Sec.~\ref{sec:similarity_complementarity}). In line with \citet{plas_better_2026}, GFM embeddings are on average partially similar and complementary, from dissimilar (Scale-MAE\pair DOFA, $0.26$) to highly similar pairs (CROMA\pair TerraMind-L, $0.83$).
Both scores depend on the dataset, and similarity does not reliably predict complementarity. Dissimilar pairs can be complementary (Prithvi\pair DOFA on DynamicEarthNet, Fig.~\ref{fig:dynamicen_scores}: similarity $0.17$, complementarity $0.10$), and on Sen1Floods11 (Fig.~\ref{fig:sen1floods11_scores}), pairs with similarity $0.36$ and $0.70$ reach comparable complementarity ($0.12$--$0.16$). Yet CROMA\pair TerraMind-L is not complementary on CropTypeMapping (Fig.~\ref{fig:croptypemapping_scores}; similarity $0.77$, complementarity $-0.03$).
The complementarity score highlights promising pairs (\eg Scale-MAE\pair DOFA on HLSBurnScars, Fig.~\ref{fig:hlsburnscars_scores}) but becomes unreliable on high-dimensional latents, where ridge regression overfits and is sensitive to regularization (\eg CROMA self-pair on MADOS, Fig.~\ref{fig:mados_scores}), and it collapses to the majority class on imbalanced datasets (\eg SpaceNet7, Fig.~\ref{fig:spacenet7_scores}).
Overall, global similarity measures may miss the localized latent signals that drive complementarity, which a linear probe may not capture reliably.

\begin{table}[t]
\vspace{-4mm}
  \centering
  \caption{\textbf{PANGAEA frozen-encoder evaluation}: mIoU as mean
$\pm$ std over $3$ seeds, on nine \textsc{Pangaea} datasets. 
Our methods are trained with $\ell_{\mathrm{mse}}+\ell_{\cos}$ of Eq.~\ref{eq:mrepa}.
Best per dataset in bold, second best is underlined.
FiveBillionPixels and Biomassters are omitted, the latter also being scored by
  RMSE. The full training and evaluation setups are detailed in App.~\ref{sec:app_training_details} and ~\ref{sec:app_pangaea} respectively.
  }
  \label{tab:pangaea_frozen_upernet}
  \resizebox{\textwidth}{!}{%
  \begin{tabular}{lccccccccc|cc}
  \toprule
  model & AI4Farms & CropMap & DynEarthNet & HLSBurns & MADOS & PASTIS & Sen1Floods11 & SN7 & xView2 & Avg mIoU & Avg rank \\
  \midrule
  UNet (baseline) & $\mathbf{45.41{\scriptstyle\,\pm\,0.71}}$ & $45.95{\scriptstyle\,\pm\,2.11}$ & $33.32{\scriptstyle\,\pm\,1.96}$ & $\mathbf{85.64{\scriptstyle\,\pm\,0.61}}$ & $57.66{\scriptstyle\,\pm\,3.00}$ & $28.56{\scriptstyle\,\pm\,0.37}$ & $\mathbf{90.88{\scriptstyle\,\pm\,0.87}}$ & $\underline{62.43{\scriptstyle\,\pm\,0.16}}$ & $42.21{\scriptstyle\,\pm\,2.32}$ & 54.67 & 4.33 \\
  \midrule
  CROMA & $\underline{27.59{\scriptstyle\,\pm\,0.27}}$ & $\mathbf{58.41{\scriptstyle\,\pm\,4.58}}$ & $30.80{\scriptstyle\,\pm\,1.07}$ & $82.59{\scriptstyle\,\pm\,0.24}$ & $\underline{65.63{\scriptstyle\,\pm\,1.58}}$ & $38.49{\scriptstyle\,\pm\,0.69}$ & $90.06{\scriptstyle\,\pm\,0.24}$ & $59.22{\scriptstyle\,\pm\,0.28}$ & $40.52{\scriptstyle\,\pm\,0.20}$ & 54.81 & 5.22 \\
  DOFA & $27.11{\scriptstyle\,\pm\,0.37}$ & $47.53{\scriptstyle\,\pm\,5.02}$ & $\underline{38.99{\scriptstyle\,\pm\,1.11}}$ & $80.86{\scriptstyle\,\pm\,0.45}$ & $60.56{\scriptstyle\,\pm\,2.64}$ & $29.53{\scriptstyle\,\pm\,0.60}$ & $88.22{\scriptstyle\,\pm\,0.20}$ & $61.01{\scriptstyle\,\pm\,0.13}$ & $59.15{\scriptstyle\,\pm\,0.10}$ & 54.77 & 5.33 \\
  Prithvi & $26.87{\scriptstyle\,\pm\,0.50}$ & $44.91{\scriptstyle\,\pm\,2.43}$ & $21.64{\scriptstyle\,\pm\,0.82}$ & $83.08{\scriptstyle\,\pm\,0.36}$ & $55.02{\scriptstyle\,\pm\,1.48}$ & $33.93{\scriptstyle\,\pm\,0.45}$ & $90.20{\scriptstyle\,\pm\,0.44}$ & $56.56{\scriptstyle\,\pm\,0.28}$ & $49.17{\scriptstyle\,\pm\,0.24}$ & 51.26 & 7.33 \\
  Scale-MAE & $22.52{\scriptstyle\,\pm\,1.87}$ & $25.18{\scriptstyle\,\pm\,1.87}$ & $35.97{\scriptstyle\,\pm\,1.75}$ & $76.56{\scriptstyle\,\pm\,0.07}$ & $55.60{\scriptstyle\,\pm\,2.25}$ & $24.61{\scriptstyle\,\pm\,0.55}$ & $74.85{\scriptstyle\,\pm\,0.09}$ & $\mathbf{62.58{\scriptstyle\,\pm\,0.50}}$ & $\mathbf{60.05{\scriptstyle\,\pm\,0.16}}$ & 48.66 & 6.67 \\
  TerraMind-L & $26.98{\scriptstyle\,\pm\,0.54}$ & $\underline{54.67{\scriptstyle\,\pm\,2.82}}$ & $35.22{\scriptstyle\,\pm\,1.42}$ & $83.41{\scriptstyle\,\pm\,0.15}$ & $\mathbf{70.99{\scriptstyle\,\pm\,2.31}}$ & $\mathbf{42.54{\scriptstyle\,\pm\,0.44}}$ & $\underline{90.78{\scriptstyle\,\pm\,0.13}}$ & $60.22{\scriptstyle\,\pm\,0.70}$ & $48.67{\scriptstyle\,\pm\,0.37}$ & \textbf{57.05} & \textbf{3.89} \\
  \midrule
  \textsc{GeoMixture-Naive} & $27.05{\scriptstyle\,\pm\,0.36}$ & $50.31{\scriptstyle\,\pm\,0.96}$ & $\mathbf{40.14{\scriptstyle\,\pm\,0.95}}$ & $83.21{\scriptstyle\,\pm\,0.14}$ & $56.55{\scriptstyle\,\pm\,4.79}$ & $38.12{\scriptstyle\,\pm\,0.55}$ & $90.27{\scriptstyle\,\pm\,0.21}$ & $61.62{\scriptstyle\,\pm\,0.37}$ & $59.20{\scriptstyle\,\pm\,0.36}$ & 56.27 & \underline{4.00} \\
  \textsc{GeoMixture-Sum} & $27.05{\scriptstyle\,\pm\,0.15}$ & $53.62{\scriptstyle\,\pm\,2.86}$ & $38.22{\scriptstyle\,\pm\,1.00}$ & $83.20{\scriptstyle\,\pm\,0.66}$ & $56.49{\scriptstyle\,\pm\,0.91}$ & $\underline{38.51{\scriptstyle\,\pm\,0.21}}$ & $90.25{\scriptstyle\,\pm\,0.28}$ & $61.72{\scriptstyle\,\pm\,0.19}$ & $59.31{\scriptstyle\,\pm\,0.30}$ & \underline{56.49} & \textbf{3.89} \\
  \textsc{GeoMixture-MoE} & $26.95{\scriptstyle\,\pm\,0.49}$ & $49.79{\scriptstyle\,\pm\,1.12}$ & $37.77{\scriptstyle\,\pm\,0.06}$ & $\underline{83.49{\scriptstyle\,\pm\,0.14}}$ & $57.26{\scriptstyle\,\pm\,2.25}$ & $38.04{\scriptstyle\,\pm\,0.32}$ & $90.10{\scriptstyle\,\pm\,0.31}$ & $61.83{\scriptstyle\,\pm\,0.08}$ & $\underline{59.39{\scriptstyle\,\pm\,0.28}}$ & 56.07 & 4.33 \\
  \bottomrule
  \end{tabular}
  }
\end{table}



\paragraph{\textsc{Pangaea} evaluation.}
We follow the \textsc{Pangaea} protocol (App.~\ref{sec:app_pangaea}) with $3$ seeds (Tab.~\ref{tab:pangaea_frozen_upernet}): encoders are frozen with a supervised UPerNet decoder, and the UNet is trained end-to-end. \textsc{GeoMixture} is aligned on the unlabeled training images of each \textsc{Pangaea-Latents} dataset, unseen by the GFMs during pretraining.
\textsc{GeoMixture-Sum} ties TerraMind-L for the best average rank ($3.89$), and all variants outperform four of the five GFMs. TerraMind-L keeps the best average mIoU ($57.05$ vs.\ $56.49$) only thanks to MADOS, without which all variants surpass it.
No model dominates, but \textsc{GeoMixture} is more consistent: \textsc{GeoMixture-Naive} is never ranked below sixth, whereas TerraMind-L ranks first or second on four datasets but sixth or seventh on four others.
\textsc{GeoMixture} trails most where one GFM clearly leads the others, \ie on MADOS, PASTIS and CropTypeMapping ($-4.0$ to $-14.5$ mIoU w.r.t.\ the best GFM), where aligning to a consensus is most costly.
Experts yield the best variant, \textsc{GeoMixture-Sum} (no trained gate), but learning the gate (\textsc{GeoMixture-MoE}) brings no further gain, and variants mostly remain within one standard deviation.
The objective matters as shown by our loss ablation study (App.~\ref{sec:app_ablation_study}, Tab.~\ref{tab:latent_moe_ablation}): $\ell_{\cos}$ is best on MADOS but worst on HLSBurnScars for every variant, and $\ell_{\mathrm{mse}}+\ell_{\cos}$ gives every variant its best average rank.

\paragraph{Frozen GFMs and a supervised baseline.}
With fewer parameters than the UPerNet decoder alone, the UNet wins three datasets and outranks four of the five GFMs, partly through AI4SmallFarms ($45.41$ vs.\ at most $27.59$ mIoU for frozen encoders). Frozen GFMs thus do not yet consistently surpass supervised training, but all \textsc{GeoMixture} variants exceed the UNet in average mIoU.

\section{Conclusion and future work}
\label{sec:conclusion}

We introduced \textsc{Pangaea-Latents}, an extension of \textsc{Pangaea} supporting research on GFM representation alignment, and \textsc{GeoMixture}, aligning a single ViT to five GFMs through a multi-representation alignment objective. 
Aligned to several teachers, \textsc{GeoMixture} ties the best GFM in average rank and is more consistent across datasets, while ranking first on only one.
Two limitations bound these results. Alignment is performed per dataset, so the student sees the target imagery, without labels, while the GFMs do not; the comparison measures how well a representation can be specialized rather than how well it transfers. The evaluation also covers nine of eleven datasets, and the alignment loss, which is not a neutral choice, is ablated on only five.
In future work, we will complete the benchmark, explore additional losses, and scale to additional GFMs \cite{hong_spectralgpt_2024, tseng_galileo_2025, astruc_anysat_2025, houdre_ramen_2025, klemmer_satclip_2025, tseng_olmoearth_2026, feng_tessera_2026}, including AlphaEarth \cite{brown_alphaearth_2025}, which releases embeddings but no backbone. Aligning on a large multimodal corpus \cite{tseng_galileo_2025} rather than per dataset would remove the specialization confound and test whether a single \textsc{GeoMixture} encoder transfers across the benchmark.

\begin{ack}
This project was supported thanks to funding from IVADO, the Canada CIFAR AI Chairs program, and the Sloan Fellowship program. 
We thank Mila -- Quebec AI Institute (mila.quebec) for supporting this project with compute resources and technical help.
\end{ack}

\bibliographystyle{unsrtnat}
\bibliography{neurips2026_roe_workshop}


\appendix

\newpage
\section{Similarity and complementarity metrics}
\label{sec:app_similarity_complementarity}

\paragraph{Similarity scores.}
Let $e_{i,k}=X \in \mathbb{R}^{n \times d_k}$ and $e_{i,l}=Y \in \mathbb{R}^{n \times d_l}$ be two GFM embeddings for a dataset $i \in \{1,\cdots,D\}$. 
We first quantified latent similarities with canonical correlation analysis (CCA) used for neural network representation similarities \cite{kornblith_similarity_2019},
\begin{equation}
  R^{2}_{\mathrm{CCA}}(X, Y) =
  \frac{\lVert Q_X^{\top} Q_Y \rVert_F^{2}}{\min(d_k, d_l)},
  \label{eq:cca}
\end{equation}
where $Q_X=X(X^\top X)^{-1/2}$ and $Q_Y=Y(Y^\top Y)^{-1/2}$ their orthonormal bases, and $\lVert \cdot \rVert_F$ the Frobenius norm.

We considered the linear centered kernel alignment (CKA) used to compare multivariate features \cite{kornblith_similarity_2019},
\begin{equation}
  \mathrm{CKA}(X, Y) =
  \frac{\lVert Y^{\top} X\rVert_F^{2}}
       {\lVert X^{\top} X \rVert_F \, \lVert Y^{\top} Y \rVert_F}.
  \label{eq:cka}
\end{equation}
where $X$ and $Y$ are centered towards the column-wise average. 
The standardized CKA is defined in Sec.~\ref{sec:similarity_complementarity} (Eq.~\ref{eq:cka_std}).

\paragraph{Complementarity score.}
Following \citet{plas_better_2026}, we measure the complementarity of two embeddings $X$ and $Y$ as
\begin{equation}
  \mathrm{compl}(X,Y) = \frac{S(X \oplus Y) - S^{\star}}{S_{\text{best}} - S^{\star}},
  \label{eq:complementarity}
\end{equation}
where $\oplus$ denotes concatenation along the feature dimension, $S(\cdot)$ is the score of a probe trained on the given embeddings, $S^{\star}$ is the score of the better of the two individual probes ($\max(S(X),S(Y))$ for higher-is-better metrics such as F1, $\min$ for lower-is-better ones such as RMSE), and $S_{\text{best}}$ is the best attainable value of the metric (\eg $1$ for F1, $0$ for RMSE).
The score is the fraction of the remaining headroom recovered by concatenation of the embeddings: it is positive when combining $X$ and $Y$ improves over the better individual probe, reaches $1$ when the combined probe attains $S_{\text{best}}$, and is negative when concatenation degrades performance.
All probes are ridge regressions whose regularization strength is selected by 4-fold cross-validation in $\{10^{-3}, 10^{-2}, 10^{-1}, 1, 10, 100\}$, scaled by the training-set size, per GFM pair and per dataset. Even so, the probe on concatenated embeddings can overfit high-dimensional latents (\eg MADOS, Fig.~\ref{fig:mados_scores}), making the score sensitive to this choice.

\section{The \textsc{Pangaea} benchmark}
\label{sec:app_pangaea}

%
%

We evaluate on \textsc{Pangaea} ~\cite{marsocci_pangaea_2025}, a standardized benchmark for geospatial
foundation models (GFMs) spanning eleven downstream datasets, four sensor families, three task
types, ground sampling distances from sub-metre to 30\,m, and single-date, bi-temporal
and time-series inputs. 
We use the released dataset splits (except for the deviations listed in Sec.~\ref{app:pangaea_deviations}),
preprocessing, decoders, optimization and metrics.
Section~\ref{app:pangaea_datasets} describes each dataset,
Section~\ref{app:pangaea_protocol} states which evaluation recipe applies to which setup, and
Section~\ref{app:pangaea_deviations} lists our deviations.

\subsection{Datasets}
\label{app:pangaea_datasets}

Table~\ref{tab:pangaea-datasets} summarises the eleven datasets. Sample counts are the lengths of
the benchmark's own loaders on our copy of the data, measured per split; these are the
quantities that are used for training and evaluation, which for two datasets differs from the
number of distinct scenes (see the table footnotes).

\begin{table}[t]
\centering
\caption{The eleven \textsc{Pangaea}  datasets. \emph{Mod.} is the modality set returned to the encoder
(O~=~optical, S~=~SAR); \emph{$T$} the number of time steps the loader returns; \emph{$C$} the number
of label classes as declared by the benchmark ($C=1$ denotes a single-logit binary mask, or, for
BioMassters, a continuous target). \emph{Size} is the scene size the loader yields. Splits are
train/val/test sample counts.}
\label{tab:pangaea-datasets}
\resizebox{\textwidth}{!}{%
\begin{tabular}{llllrrrlr}
\toprule
Dataset & Sensor(s) & Domain / task & Mod. & GSD (m)\footnotemark[1] & Size & $T$ & $C$ & Splits \\
\midrule
MADOS             & Sentinel-2            & Marine pollution, seg.   & O    & 10          & $240^2$  & 1  & 15 & 1433/642/728 \\
HLSBurnScars      & HLS (L8\,+\,S2)       & Wildfire, seg.           & O    & 30          & $512^2$  & 1  & 2  & 486/54/264 \\
Sen1Floods11      & Sentinel-1/2          & Flood, seg.              & O+S  & 10          & $512^2$  & 1  & 2  & 252/89/90 \\
AI4SmallFarms     & Sentinel-2            & Field boundaries, seg.   & O    & 10          & $496^2$  & 1  & 1  & 43/9/10 \\
CropTypeMapping   & Sentinel-1/2          & Crop type, seg.          & O+S  & 10          & $64^2$   & 6  & 4  & 554/66/67 \\
PASTIS          & Sentinel-1/2          & Crop type, seg.          & O+S  & 10          & $128^2$  & 6  & 20 & 1455/482/496 \\
DynamicEarthNet   & PlanetFusion          & Land cover, seg.         & O    & 3           & $1024^2$ & 6  & 6  & 936/192/192 \\
FiveBillionPixels & Gaofen-2              & Land cover, seg.         & O    & 4           & $520^2$  & 1  & 25 & 19110/4004/4186 \\
SpaceNet7        & Planet                & Urban growth, CD         & O    & $\approx4$  & $256^2$  & 2  & 1  & 480/240/224 \\
xView2 (xBD)      & Maxar                 & Building damage, CD      & O    & $\leq0.8$   & $1024^2$ & 2  & 5  & 13384/917/933\\
BioMassters       & Sentinel-1/2          & Biomass, regression      & O+S  & 10          & $256^2$  & 12 & 1  & 7820/869/2773 \\
\bottomrule
\end{tabular}}
\end{table}
\footnotetext[1]{GSD is that of the source product, taken from each dataset's own publication; it is
not declared in the benchmark's configuration files.}

\paragraph{MADOS.}
Marine Debris and Oil Spill detection from Sentinel-2. Each sample is a $240\times240$ single-date
patch of ACOLITE-processed Rayleigh-corrected reflectance, with eleven spectral bands (B1--B8, B8A,
B11, B12) resampled to a 10\,m grid, and a 15-class label map covering floating
litter, algal and organic slicks, oil spills, ships and platforms, and several water types. The
annotation is sparse where most pixels carry the ignore index, so the benchmark crops around
labelled pixels rather than uniformly at random (Sec.~\ref{app:pangaea_protocol}). Classes are
strongly imbalanced: marine water, sediment-laden water and oil spill account for $74\%$ of
labelled pixels, while marine debris, foam and jellyfish each fall below $1\%$. The dataset probes
fine-grained spectral discrimination of globally distributed, low-prevalence targets.

\paragraph{HLSBurnScars.}
Burn-scar delineation on Harmonized Landsat and Sentinel-2 (HLS) surface reflectance over the
United States. Scenes are $512\times512$ at 30\,m with six bands (B2, B3, B4, B8A, B11,
B12) and a binary label (burned / not burned, $\approx11\%$ positive). The benchmark's test split is
the release's held-out validation set; its train and validation splits are a fixed $90/10$ random
partition of the release's training set. It is the only 30\,m dataset in the benchmark and
therefore the main test of transfer across ground sampling distance.

\paragraph{Sen1Floods11.}
Flood-water mapping from co-registered Sentinel-1 GRD and Sentinel-2 L1C acquisitions of eleven
flood events worldwide. The benchmark uses the hand-labelled subset: $512\times512$ scenes with 13
optical bands and two SAR polarisations (VV, VH), and a binary water mask ($\approx9.5\%$ positive).
Because SAR penetrates cloud while optical does not, it is the benchmark's cleanest measure of
multi-modal fusion; an encoder can be evaluated on either modality alone or on both.

\paragraph{AI4SmallFarms.}
Smallholder field-boundary delineation in Cambodia and Vietnam from Sentinel-2 (B2, B3, B4, B8),
$496\times496$ scenes, a single binary crop-field mask ($\approx27\%$ positive). With 62 scenes in
total it is by far the smallest dataset in the benchmark and the one most sensitive to the quality of
the frozen representation, since the decoder has very little data from which to recover. It is the
only dataset for which the benchmark replicates the data loader, visiting each train and validation
scene twice per epoch.

\paragraph{CropTypeMapping (South Sudan).}
Crop-type segmentation from paired Sentinel-1 and Sentinel-2 time series over South Sudan, using the
release's official partition. Each sample is a $64\times64$ scene with ten optical bands and two SAR
polarisations; the acquisition series is truncated or zero-padded to exactly $T=6$ steps. Only the
four most frequent crop types (sorghum, maize, rice, groundnut) are retained, every other pixel is
mapped to the ignore index, and sorghum alone covers $68\%$ of the labelled pixels. Although the
benchmark's documentation lists Planet among the sensors, its loader returns only the Sentinel-1 and
Sentinel-2 arrays. The scenes are far smaller than any encoder's input resolution, so they are
resized rather than cropped, and evaluated whole rather than by sliding window. It combines  multi-temporal and multi-modal axes of the benchmark at the smallest spatial support.

\paragraph{PASTIS.}
Agricultural parcel classification in France from Sentinel-2 and ascending-orbit Sentinel-1 time
series, using the official five-fold partition (folds 1-3 train, 4 validation, 5 test). Samples are
$128\times128$ patches with ten optical bands and three SAR channels (VV, VH, VV$-$VH); $T=6$ frames
are drawn evenly spaced across each patch's acquisition series. The label space has 20 entries: a
background class, 18 crop classes, and a void label that serves as the ignore index and itself covers
$29\%$ of pixels. Class frequencies among the evaluated classes span nearly two orders of magnitude,
from meadow ($26\%$) to potatoes ($0.5\%$). The loader also opens each patch's SPOT-6 aerial tile,
which must therefore be present on disk, but does not return it. PASTIS is the benchmark reference
test of temporal modelling: crop types are separable mainly through their phenological trajectory,
not through any single date.

\paragraph{DynamicEarthNet.}
Land-cover mapping from PlanetFusion daily surface-reflectance mosaics at 3\,m. A sample is
one labelled month of one $1024\times1024$ cube: the benchmark feeds the first six \emph{daily}
mosaics of that month ($T=6$, four bands: blue, green, red, NIR) and predicts the month's
six-class land-cover map (impervious surface, agriculture, forest and other vegetation, wetlands,
soil, water); the release's seventh label band, snow and ice, becomes the ignore index. Because
DynamicEarthNet withheld its official validation and test labels for its challenge, any usable
partition divides the 55 publicly labelled cubes; the benchmark uses a published cube-level split
($39/8/8$ cubes over 24 labelled months each). Wetlands is a $\approx0.3\%$ class in our
measurements, the value published in the benchmark's configuration ($7\%$) is inconsistent with
the data, so its per-class IoU is noisy and depresses mIoU independently of model quality;
per-class IoU is the more informative comparison on this dataset.

\paragraph{FiveBillionPixels.}
Urban and peri-urban land cover in China from Gaofen-2 imagery at 4\,m, four bands in
NIR-R-G-B order. The benchmark evaluates non-overlapping $520\times520$ crops of the original
large scenes, with 25 land-cover classes of which the \emph{unlabeled} class is the ignore index. At
27{,}300 crops it is the largest segmentation dataset and, with 24 evaluated classes, the one
with the finest-grained label space.

\paragraph{SpaceNet 7.}
Urban development from Planet monthly mosaics at $\approx$4\,m, globally distributed. The
benchmark evaluates this dataset as binary change detection: each $1024\times1024$ scene is tiled
into sixteen $256\times256$ tiles, two acquisitions of a tile are encoded, and a single change mask
is predicted, the mask being the pixelwise difference between the building-footprint rasters of
the two dates. Train, validation and test are 30, 15 and 14 disjoint areas of interest, so the split
is geographic. The temporal pair is drawn differently per split: training samples a random pair at
least five acquisitions apart, while validation and test use a fixed, deterministic pairing.

\paragraph{xView2.}
Building-damage assessment from sub-metre Maxar RGB imagery over disasters worldwide. Each sample
stacks a pre-event and a post-event acquisition into a $(3, 2, 1024, 1024)$ tensor; the target is the
post-event mask with five classes: no building, and four ordinal damage levels (no damage, minor,
major, destroyed). Unlike SpaceNet7 the change is \emph{typed} rather than binary, so the paired
representation must carry damage severity, not merely difference. The benchmark oversamples training
scenes containing damage, which is why the training loader is longer than the number of distinct
pairs. 

\paragraph{BioMassters.}
Above-ground biomass regression over Finnish forests from Sentinel-1 and Sentinel-2. Each sample is a
$256\times256$ scene with $T=12$ monthly composites, ten optical bands plus a Sentinel-2
cloud-probability channel, and four SAR channels (ascending and descending VV/VH); months with no
acquisition are filled with zeros. The target is continuous, so the benchmark trains with MSE, reports
RMSE, and normalizes inputs by published minima and maxima rather than by mean and standard
deviation. It is the only regression task, the only 12-step series, and the only dataset whose metric
is lower-is-better.

\subsection{Evaluation protocol}
\label{app:pangaea_protocol}

\textsc{Pangaea}  evaluates a foundation model as a \emph{frozen} feature extractor: the encoder's weights are
never updated and only a task decoder is trained on its multi-scale features. The recipe is
determined by three properties of the (encoder, dataset) pair, the task type (whether the loader
returns a time series), and which modalities the two share, and is otherwise identical across the
grid. Table~\ref{tab:pangaea-recipe} gives the mapping.

\begin{table}[t]
\centering
\caption{Which evaluation components are used for which setup. $T$ is the number of time steps
returned by the loader. The from-scratch UNet baseline is the only model trained end-to-end.}
\label{tab:pangaea-recipe}
\resizebox{\textwidth}{!}{%
\begin{tabular}{llll}
\toprule
Task & Temporality & Decoder (frozen GFM) & Loss \\
\midrule
Segmentation      & $T=1$ & UPerNet                         & Cross-entropy ($C>1$) / Dice ($C=1$) \\
Segmentation      & $T>1$ & UPerNet + L-TAE temporal fusion & Cross-entropy \\
Change detection  & $T=2$ & Siamese-concatenation UPerNet   & Cross-entropy ($C>1$) / Dice ($C=1$) \\
Regression        & $T=1$ & Regression UPerNet              & MSE \\
Regression        & $T>1$ & Regression UPerNet + L-TAE      & MSE \\
\midrule
\multicolumn{4}{l}{\emph{UNet baseline (trained from scratch):} UNet / Siamese UNet / regression
UNet; time steps folded into channels.} \\
\bottomrule
\end{tabular}
}
\end{table}

\paragraph{Decoders.}
Single-date segmentation uses a UPerNet head with 512 channels over the encoder's feature pyramid.
For multi-temporal datasets (CropTypeMapping, PASTIS, DynamicEarthNet, BioMassters) the frozen
encoder is applied independently to each time step and the resulting sequence of feature maps is
fused by a lightweight temporal attention encoder (L-TAE) before the UPerNet head; this is what
allows single-date foundation models to be evaluated on time series at all. Change detection
(SpaceNet7, xView2) uses a Siamese UPerNet that encodes both acquisitions with the same frozen
weights and concatenates their features before decoding. Regression replaces the classification head
with a regression UPerNet. The from-scratch UNet baseline brings its own decoder family and, being
a convolutional network without a multi-scale token grid, handles time by folding the temporal axis
into the input channels. In every case the decoder's logits are bilinearly interpolated to the
target's spatial size before the loss and the metrics are computed.

\paragraph{Input adaptation.}
Every scene is reduced to the encoder's expected input (224\,px for DOFA, Prithvi, Scale-MAE
and TerraMind, 120\,px for CROMA), and the native scene size for UNet by one of three
policies. Scenes larger than the encoder input are randomly cropped at training time and processed
by sliding window at test time. MADOS instead uses a \emph{focus crop} guaranteed to contain labelled
pixels, because a uniform crop of its sparse annotation frequently yields an all-ignore target and an
undefined loss. Scenes smaller than the encoder input (CropTypeMapping at $64^2$, PASTIS at
$128^2$) are bilinearly resized rather than cropped (the label is left at its native resolution)
and are evaluated whole rather than by sliding window. Spectral bands are then filtered to those the
encoder declares, normalized, and zero-padded where the encoder expects a band the dataset does not
provide, so an encoder is never silently fed a mismatched channel order. Normalization is per-band
mean/standard-deviation for segmentation and change detection and min--max for regression. For
modality-flexible encoders the modality set is restricted to the intersection with the dataset's, so
an any-to-any model is evaluated on optical-only data rather than being excluded or fed a fake
SAR input.

\paragraph{Optimization.}
All runs use AdamW ($\text{lr}=10^{-4}$, $\beta=(0.9,0.999)$, weight decay $0.05$) with a multi-step
schedule that decays the learning rate at $60\%$ and $90\%$ of the total iterations, batch size 8 per
device, fp32 precision, and 80 epochs.
The loss is cross-entropy for multi-class tasks, Dice for the single-logit binary masks
(AI4SmallFarms, SpaceNet7), and MSE for regression. Validation runs every 5 epochs and the reported
checkpoint is the best-scoring one: by validation mIoU for every segmentation and change-detection
dataset except AI4SmallFarms, which selects on the IoU of its positive class, and by validation MSE
for regression.

\paragraph{Inference and metrics.}
Segmentation and change detection are reported as mIoU, the mean of the per-class IoU over all
classes other than the ignore index; the binary datasets are scored over two classes, since the
benchmark's evaluator sizes its confusion matrix from the class-name list even where the decoder
emits a single logit. Regression is reported as RMSE. Every (encoder, dataset) cell is run with three
seeds and we report mean $\pm$ standard deviation. Because RMSE is lower-is-better while mIoU is
higher-is-better, across datasets with different metrics, aggregates are computed as average rank; over the nine mIoU datasets of Tab.~\ref{tab:pangaea_frozen_upernet}, we also report average mIoU.

\subsection{Deviations from the published protocol}
\label{app:pangaea_deviations}

We record every deviation from the \textsc{Pangaea} protocol, since each one breaks comparability with the
published numbers if left implicit.

\begin{itemize}
  \item \textbf{SpaceNet7 is evaluated as change detection.} The single-date mapping loader returns
  a rank-3 tensor while the preprocessing pipeline asserts rank~4, and fails at every upstream
  revision. The change-detection loader is unaffected, and is also the configuration \textsc{Pangaea} reports
  for this dataset.
  \item \textbf{The FiveBillionPixels tiles and their partition are ours.} The benchmark evaluates
  non-overlapping $520\times520$ crops of the original Gaofen-2 scenes, but that cropped release was
  never published. We derive it from the original product (8-bit NIR--RGB scenes and index
  annotations, which is what the benchmark's band list and statistics assume) and assign whole
  scenes (not tiles) to a seeded $70/15/15$ partition, so neighbouring tiles cannot straddle a
  split.
  \item \textbf{The BioMassters chip lists are ours.} The benchmark reads per-split chip lists that were not released. We use the release's own test set and carve a seeded $10\%$ validation split out of
  its training set.
  \item \textbf{The cloud-probability channel is dropped for DOFA on BioMassters.} BioMassters lists
  Sentinel-2's cloud-probability mask as an eleventh ``optical band''. Encoders with a fixed band list
  never request it; DOFA, which looks up a centre wavelength for every requested band, has no entry
  for a mask. We omit the channel rather than assign it a fictitious wavelength. It is the last band,
  so the remaining channel indices are unchanged.
  \item \textbf{Reduced batch size for the UNet baseline at $1024^2$.} UNet retains full-resolution
  feature maps and exhausts GPU memory on xView2 and DynamicEarthNet at the default batch size; those
  two cells use batch size 2. The transformer encoders operate on a $16\times$-downsampled grid and
  are unaffected.
  \item \textbf{Deterministic temporal pairing for SpaceNet7 feature caching.} The change-detection
  loader samples its two timestamps at random in training mode. Where we cache encoder features, the
  loader is placed in its deterministic evaluation pairing so that a cached feature is a function of
  the sample index. Benchmark training runs are unaffected.
\end{itemize}

Two further properties are upstream behaviour rather than deviations, but affect how the
corresponding numbers should be read. DynamicEarthNet's published class distribution is internally
inconsistent, and its wetlands class is $\approx0.3\%$ rather than the stated $7\%$. And
CropTypeMapping ships one corrupt archive, which the loader handles by returning the following sample
in its place, so one scene of 554 is duplicated in any full pass over that split.

\section{Description of \textsc{Pangaea-Latents}}
\label{sec:app_pangaea_latents}


%

Aligning a student to several GFMs requires their embeddings of the
same input. Recomputing them every epoch is wasteful and, for encoders that must be held
frozen, unnecessary. We therefore precompute them once into \textsc{Pangaea-Latents} (see Sec.~\ref{sec:pangaea_latents}).
Because the tiling is shared, a given (sample, tile) index refers to the same
spatial region for every GFM, which is what makes the embeddings comparable and the alignment target well defined.

\subsection{What is stored}

A record is the last-layer grid of one tile of shape $(d, T, H', W')$ stored in \texttt{bfloat16}. The
time axis is always kept and never pooled; no pooled vector is stored, since the projection into a
shared space is learned downstream. 
For our experiments with multi-temporal datasets, alignment is applied per frame.
Table~\ref{tab:latents-fm} gives the per-FM geometry.

\begin{table}[t]
\centering
\caption{The five GFMs and the grid each produces per tile. \emph{Input} is the model
own window, to which every shared tile is resized; the token grid is $(\text{input}/\text{patch})^2$.}
\label{tab:latents-fm}
\begin{tabular}{lrrrl}
\toprule
Model & $d$ & Input & Patch & Grid \\
\midrule
CROMA (optical)  & 1024 & 120 & 8  & $15\times15$ \\
DOFA             & 768  & 224 & 16 & $14\times14$ \\
Prithvi          & 768  & 224 & 16 & $14\times14$ \\
Scale-MAE         & 1024 & 224 & 16 & $14\times14$ \\
TerraMind-L& 1024 & 224 & 16 & $14\times14$ \\
\bottomrule
\end{tabular}
\end{table}

Each (dataset, GFM, split) set is written as fixed-shape sharded \texttt{.npy} arrays (512 tiles per
shard) with a structured index recording, per row, the source sample index and the tile's top-left
pixel. \texttt{bfloat16} has no NumPy dtype, so grids are stored bit-exact through a
\texttt{uint16} view. Every set carries a recording of the \textsc{Pangaea}  pin, the dataset version,
the encoder weight digest, the code revision, and content hashes of both the latent specification
and the tiling policy.
CROMA grids are bilinearly resized to $14\times14$ for all our experiements to unify the grid size over all GFMs.

\subsection{The shared tiling}

Tiles are cut at native resolution on a shared $224$-pixel grid (the largest GFM window) at
positions laid out by the same rule as \textsc{Pangaea} sliding-window evaluator, evenly spaced over
$[0,\,\text{size}-\text{tile}]$. A GFM with a smaller window (CROMA, 120) resizes each tile
internally. The tile set is deliberately \emph{not} identical to the evaluator's: the evaluator
opens $\lceil \text{size}/\text{input}\rceil$ windows at each encoder's own input size, which
would give a different set of crops per GFM, whereas a shared index requires one grid common to
all five.
Three policies cover the eleven datasets (Table~\ref{tab:latents-view}): a $3\times3$ grid for
scenes much larger than a tile; a single centre tile for scenes only marginally larger, where a grid
would be almost entirely redundant overlap; and a single whole-scene tile, at the GFM's input size,
for the three datasets \textsc{Pangaea}  itself feeds whole.

\begin{table}[t]
\centering
\caption{Tiling policy, subsampling cap and stored tiles per dataset. Caps are on \emph{samples} per
split, drawn with a fixed seed; the four capped datasets would otherwise dominate the dataset.
Tiles are summed over train/val/test and are identical for all five GFMs.}
\label{tab:latents-view}
\begin{tabular}{llllr}
\toprule
Dataset & Scene & $T$ & Policy & Tiles \\
\midrule
MADOS             & $240^2$  & 1  & centre           & 2803 \\
HLSBurnScars      & $512^2$  & 1  & $3\times3$ grid  & 7236 \\
Sen1Floods11      & $512^2$  & 1  & $3\times3$ grid  & 3879 \\
AI4SmallFarms     & $496^2$  & 1  & $3\times3$ grid  & 558 \\
CropTypeMapping   & $64^2$   & 6  & whole scene      & 687 \\
PASTIS          & $128^2$  & 6  & whole scene      & 2433 \\
DynamicEarthNet   & $1024^2$ & 6  & $3\times3$ grid, cap 220/40/40   & 2700 \\
FiveBillionPixels & $520^2$  & 1  & $3\times3$ grid, cap 1200/200/200 & 14400 \\
SpaceNet7        & $256^2$  & 2  & centre           & 944 \\
xView2            & $1024^2$ & 2  & $3\times3$ grid, cap 600/100/100 & 7200 \\
BioMassters       & $256^2$  & 12 & whole scene, cap 1200/200/200    & 1600 \\
\bottomrule
\end{tabular}
\end{table}

All $5\times11=55$ cells are built, every GFM shares the optical modality with every dataset. The
dataset holds $44{,}440$ tiles per GFM, $222{,}200$ across the 55 cells, which is $496{,}420$
per-frame embeddings once the time axis is counted, for $185.2$\,GB in \texttt{bfloat16}.

\subsection{Extraction}

One sample is processed as follows. It is read through \textsc{Pangaea}'s \emph{validation} preprocessor
for that dataset: band filtering, the dataset's published normalization (mean/standard deviation,
or min--max for the regression task), band padding, and, for the datasets the benchmark resizes, its
resize with no random crop. Inputs are normalized exactly as the benchmark
normalizes it at evaluation time. The normalized scene is then tiled by the policy above, each tile is resized to the
GFM's window, and the frozen encoder's last-layer output is written to the dataset.

Two details are related to the time axis. An encoder that natively uses time series as input (Prithvi) receives all
$T$ frames in one forward pass, but only where the \emph{dataset configuration} declares a time
series; single-date encoders are looped frame by frame and their grids stacked back into a $T$ axis.
Either way the stored grid has the same layout, so the time axis is available downstream regardless
of which GFM produced it. Encoders are instantiated through the same code path as the benchmark
sweep, including the per-cell recipe fixes, so a modality-flexible encoder is restricted to the
dataset's modalities exactly as it is at evaluation.

Because the GFMs expect different band sets, what each one actually consumes differs: \textsc{Pangaea}'s band
filter keeps the bands an encoder requests and the padding step zero-fills any it expects but the
dataset lacks. DOFA, whose band list is the dataset's own, receives every available band unpadded;
the fixed-band encoders (CROMA and TerraMind-L at 12 optical bands, Prithvi at 6, Scale-MAE at 3) see
correspondingly less, and the optical-only encoders ignore SAR where a dataset provides it. These
padded channels carry no information and are recorded per cell rather than left implicit.

\subsection{Determinism and verification}

An embedding is only recorded if it is a function of its index. Two sources of randomness had to
be removed. \textsc{Pangaea} SpaceNet7 change-detection loader draws its temporal pair at random in
training mode; extraction forces the loader into its deterministic evaluation pairing. PASTIS
subsamples any acquisition series longer than 50 frames with an unseeded permutation, in every
split; extraction reads each sample under a seed derived from its
$(\text{dataset},\text{split},\text{index})$ identity, so the draw depends on the sample rather than
on when it was read.

Correctness is then checked by re-extraction: random tiles are re-derived through the identical
build path and compared to the dataset, and the check passes only on \emph{bitwise} equality. This
tests index correctness and forward determinism in one, and it is what caught the SpaceNet7 issue
above.

\subsection{Limitations}

The tiling is a first version and is marked provisional in the store's own manifest. Two
consequences are worth stating. Tiles are cut at $224$ pixels and resized to each GFM's window, so
CROMA, whose window is $120$, sees the shared tile at a different effective ground sampling distance
than it was pretrained at. And the tiling density is a coarse setting rather than a tuned one: the $3\times3$ grid covers
the $496$--$520$ pixel scenes completely, but on the $1024$ pixel ones (DynamicEarthNet, xView2) it
leaves $352$ of $1024$ pixels per axis unseen, where full coverage would need $5\times5$; and the
centre policy discards an $8$ pixel border on MADOS and a $16$ pixel one on SpaceNet7. Finally, the
per-sample seeding makes our extraction reproducible, but it does not make PASTIS's frame
selection match what the benchmark itself samples at training time, which remains random run to run
upstream.

\section{\textsc{GeoMixture} training details}
\label{sec:app_training_details}
The three variants differ only in the module between the backbone and the projections: none (\textsc{Naive}), summed experts (\textsc{Sum}), or routed experts (\textsc{MoE}).
The student is a ViT-B/16 initialized from ImageNet, its patch embedding inflated from RGB to the dataset's input bands and left fully trainable using $224^2$ tiles as input. 
Training is supervised with the frozen latents of the five teachers on the common $14\times14$ token grid.
Optimization is AdamW with weight decay $0.05$ and with linear warmup over $3\%$ of the epoch budget and cosine decay to $1\%$ of the peak. Batch size ($32$ to $128$), epoch budget ($70$ to $500$) and peak learning rate are set per dataset and provided in Table~\ref{tab:align-hparams}. We keep the checkpoint with the lowest validation alignment loss, scored with the same objective that is optimized, and evaluate it frozen under the Pangaea protocol over three seeds.
\textsc{Pangaea-Latents} contains all pairs of datasets and GFMs, thus we set $m=1$ in all our experiments where the mask allows partial teacher setup (Eqs.~\ref{eq:moe},~\ref{eq:mrepa}).

\begin{table}[t]
\centering
\caption{Per-dataset alignment hyperparameters for the $\ell_{\mathrm{mse}}+\ell_{\cos}$ runs,
ordered by epoch budget. \textsc{GeoMixture-Sum} and \textsc{GeoMixture-MoE} use $10^{-4}$ LR on every dataset; only
\textsc{GeoMixture-Naive} departs from it, on three. Epochs, steps, batch size and $n_{\text{train}}$ are
shared by all three variants.}
\label{tab:align-hparams}
\begin{tabular}{lccrrrr}
\toprule
 & \multicolumn{2}{c}{Learning rate (LR)} & & & & \\
\cmidrule(lr){2-3}
Dataset & \textsc{Sum}, \textsc{MoE} & \textsc{Naive} & Epochs & Steps & Batch & $n_{\text{train}}$ \\
\midrule
AI4SmallFarms   & $10^{-4}$ & $5{\times}10^{-4}$ & 500 & 6047  & 32  & 387   \\
MADOS           & $10^{-4}$ & $10^{-4}$          & 300 & 6718  & 64  & 1433  \\
SpaceNet7      & $10^{-4}$ & $5{\times}10^{-4}$ & 300 & 4500  & 64  & 960   \\
CropTypeMapping & $10^{-4}$ & $10^{-4}$          & 300 & 15582 & 64  & 3324  \\
Sen1Floods11    & $10^{-4}$ & $10^{-3}$          & 200 & 3544  & 128 & 2268  \\
HLSBurnScars    & $10^{-4}$ & $10^{-4}$          & 200 & 6835  & 128 & 4374  \\
PASTIS        & $10^{-4}$ & $10^{-4}$          & 200 & 13641 & 128 & 8730  \\
xView2          & $10^{-4}$ & $10^{-4}$          & 150 & 12657 & 128 & 10800 \\
DynamicEarthNet & $10^{-4}$ & $10^{-4}$          & 70  & 6497  & 128 & 11880 \\
\bottomrule
\end{tabular}
\end{table}

\newpage
\section{Loss ablation study for \textsc{GeoMixture}}
\label{sec:app_ablation_study}

\begin{table}[h]
  \centering
  \caption{Alignment loss ablation (\textsc{Pangaea}, UPerNet decoder): mIoU as
  mean $\pm$ std over $3$ seeds, per \textsc{GeoMixture} variant $\times$
  alignment loss $\ell$ of Eq.~\ref{eq:mrepa}. Best per dataset in bold,
  second best is underlined. Avg rank is a variant's mean
  rank across the five datasets, where rank~$1$ is the best of the nine variants
  on that dataset (lower is better). 
  Ranks are computed on unrounded values.
  The ablation covers the five datasets run
  with every loss; the remaining \textsc{Pangaea} datasets appear in
  Tab.~\ref{tab:pangaea_frozen_upernet}. Datasets acronyms AI4Farms, HLSBurns, and SN7 stands for AI4SmallFarms, HLSBurnScars, SpaceNet7 respectively.}
  \label{tab:latent_moe_ablation}
  \resizebox{\textwidth}{!}{%
  \begin{tabular}{llccccc|cc}
  \toprule
  variant & $\ell$ & AI4Farms & HLSBurns & MADOS & Sen1Floods11 & SN7 & Avg mIoU & Avg rank \\
  \midrule
  \multirow{3}{*}{\textsc{GeoMixture-Naive}} & $\ell_{\mathrm{mse}}$ & $\mathbf{27.49{\scriptstyle\,\pm\,0.41}}$ & $83.31{\scriptstyle\,\pm\,0.09}$ & $52.16{\scriptstyle\,\pm\,1.71}$ & $\mathbf{90.59{\scriptstyle\,\pm\,0.01}}$ & $61.32{\scriptstyle\,\pm\,0.36}$ & 62.98 & \underline{4.00} \\
   & $\ell_{\cos}$ & $\underline{27.31{\scriptstyle\,\pm\,0.39}}$ & $82.31{\scriptstyle\,\pm\,0.88}$ & $\underline{58.31{\scriptstyle\,\pm\,2.32}}$ & $\underline{90.48{\scriptstyle\,\pm\,0.15}}$ & $61.38{\scriptstyle\,\pm\,0.35}$ & \underline{63.96} & \underline{4.00} \\
   & $\ell_{\mathrm{mse}}+\ell_{\cos}$ & $27.05{\scriptstyle\,\pm\,0.36}$ & $83.21{\scriptstyle\,\pm\,0.14}$ & $56.55{\scriptstyle\,\pm\,4.79}$ & $90.27{\scriptstyle\,\pm\,0.21}$ & $61.62{\scriptstyle\,\pm\,0.37}$ & 63.74 & \textbf{3.80} \\
  \midrule
  \multirow{3}{*}{\textsc{GeoMixture-Sum}} & $\ell_{\mathrm{mse}}$ & $26.67{\scriptstyle\,\pm\,0.21}$ & $82.45{\scriptstyle\,\pm\,0.60}$ & $53.00{\scriptstyle\,\pm\,2.20}$ & $90.16{\scriptstyle\,\pm\,0.18}$ & $61.47{\scriptstyle\,\pm\,0.19}$ & 62.75 & 6.40 \\
   & $\ell_{\cos}$ & $27.04{\scriptstyle\,\pm\,0.22}$ & $82.36{\scriptstyle\,\pm\,1.04}$ & $58.09{\scriptstyle\,\pm\,2.20}$ & $89.78{\scriptstyle\,\pm\,0.06}$ & $61.04{\scriptstyle\,\pm\,0.54}$ & 63.66 & 6.60 \\
   & $\ell_{\mathrm{mse}}+\ell_{\cos}$ & $27.05{\scriptstyle\,\pm\,0.15}$ & $83.20{\scriptstyle\,\pm\,0.66}$ & $56.49{\scriptstyle\,\pm\,0.91}$ & $90.25{\scriptstyle\,\pm\,0.28}$ & $\underline{61.72{\scriptstyle\,\pm\,0.19}}$ & 63.74 & \underline{4.00} \\
  \midrule
  \multirow{3}{*}{\textsc{GeoMixture-MoE}} & $\ell_{\mathrm{mse}}$ & $26.98{\scriptstyle\,\pm\,0.48}$ & $\mathbf{83.73{\scriptstyle\,\pm\,0.28}}$ & $47.38{\scriptstyle\,\pm\,2.55}$ & $90.10{\scriptstyle\,\pm\,0.16}$ & $61.51{\scriptstyle\,\pm\,0.18}$ & 61.94 & 5.60 \\
   & $\ell_{\cos}$ & $27.02{\scriptstyle\,\pm\,0.17}$ & $81.95{\scriptstyle\,\pm\,0.63}$ & $\mathbf{60.10{\scriptstyle\,\pm\,2.94}}$ & $89.94{\scriptstyle\,\pm\,0.35}$ & $61.27{\scriptstyle\,\pm\,0.16}$ & \textbf{64.06} & 6.40 \\
   & $\ell_{\mathrm{mse}}+\ell_{\cos}$ & $26.95{\scriptstyle\,\pm\,0.49}$ & $\underline{83.49{\scriptstyle\,\pm\,0.14}}$ & $57.26{\scriptstyle\,\pm\,2.25}$ & $90.10{\scriptstyle\,\pm\,0.31}$ & $\mathbf{61.83{\scriptstyle\,\pm\,0.08}}$ & 63.93 & 4.20 \\
  \bottomrule
  \end{tabular}
  }
\end{table}

\section{Similarity and complementarity scores}
\label{sec:app_scores}

\begin{figure}[h]
  \centering
  \begin{subfigure}[b]{0.245\linewidth}
    \includegraphics[trim=12 38 11 46, clip, width=\linewidth]{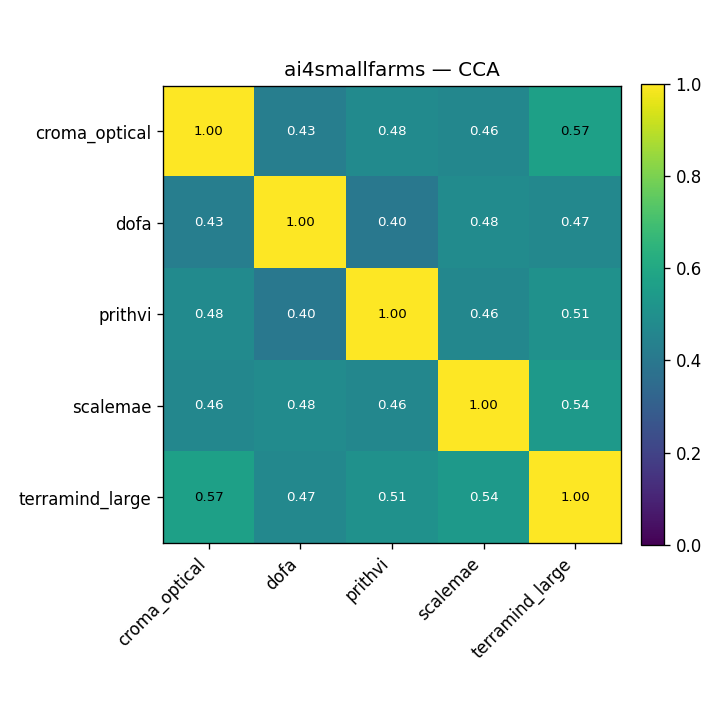}
    \caption{CCA}
    \label{fig:ai4smallfarms_scores_a}
  \end{subfigure}
  \begin{subfigure}[b]{0.245\linewidth}
    \includegraphics[trim=12 38 11 46, clip, width=\linewidth]{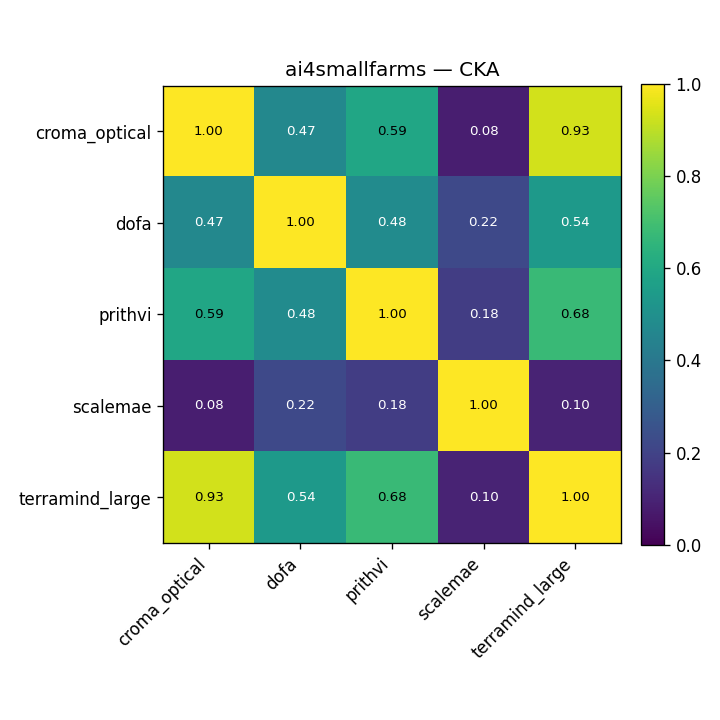}
    \caption{CKA}
    \label{fig:ai4smallfarms_scores_b}
  \end{subfigure}
  \begin{subfigure}[b]{0.245\linewidth}
    \includegraphics[trim=12 38 11 48, clip, width=\linewidth]{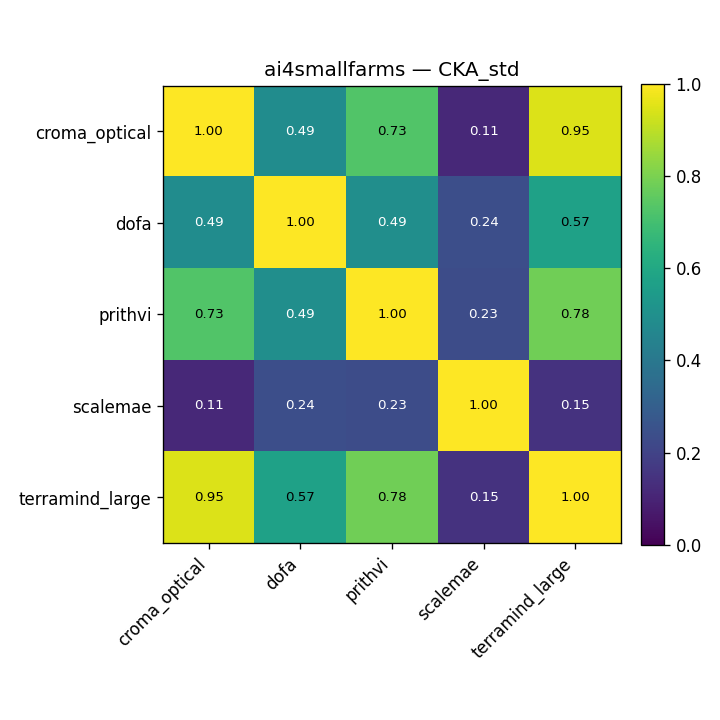}
    \caption{Standardized CKA}
    \label{fig:ai4smallfarms_scores_c}
  \end{subfigure}
  \begin{subfigure}[b]{0.245\linewidth}
    \includegraphics[trim=12 38 11 52, clip, width=\linewidth]{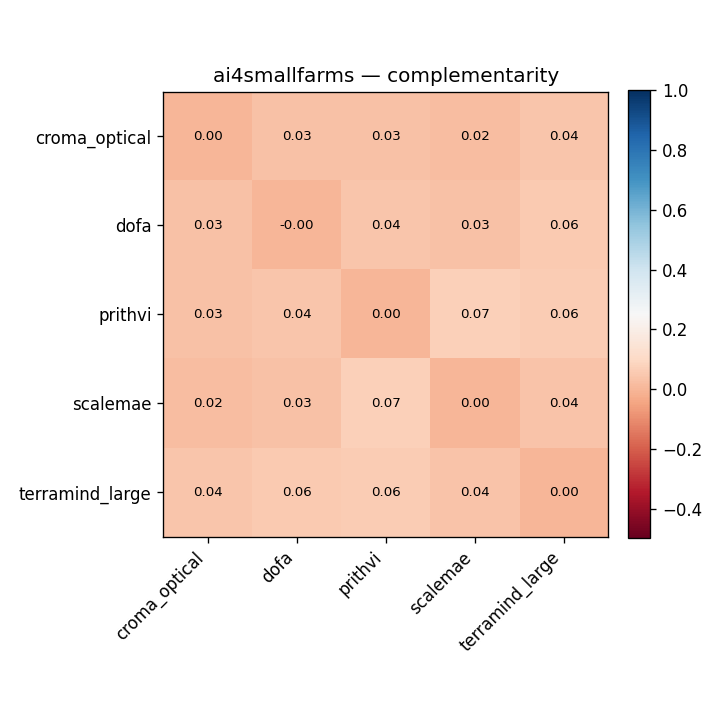}
    \caption{Complementarity}
    \label{fig:ai4smallfarms_scores_d}
  \end{subfigure}
  \caption{AI4SmallFarms similarity and complementarity scores.}
  \label{fig:ai4smallfarms_scores}
\end{figure}

\begin{figure}[h]
  \centering
  \begin{subfigure}[b]{0.245\linewidth}
    \includegraphics[trim=12 38 11 46, clip, width=\linewidth]{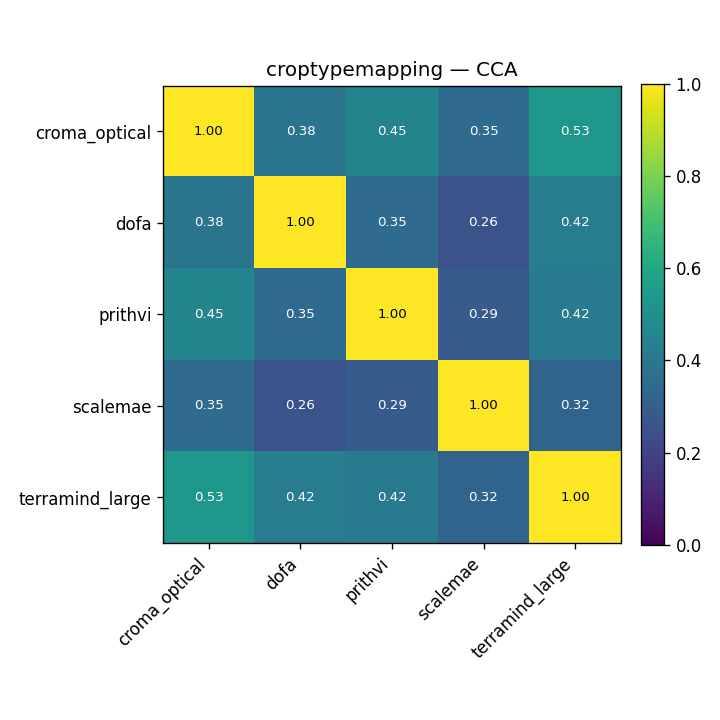}
    \caption{CCA}
    \label{fig:croptypemapping_scores_a}
  \end{subfigure}
  \begin{subfigure}[b]{0.245\linewidth}
    \includegraphics[trim=12 38 11 46, clip, width=\linewidth]{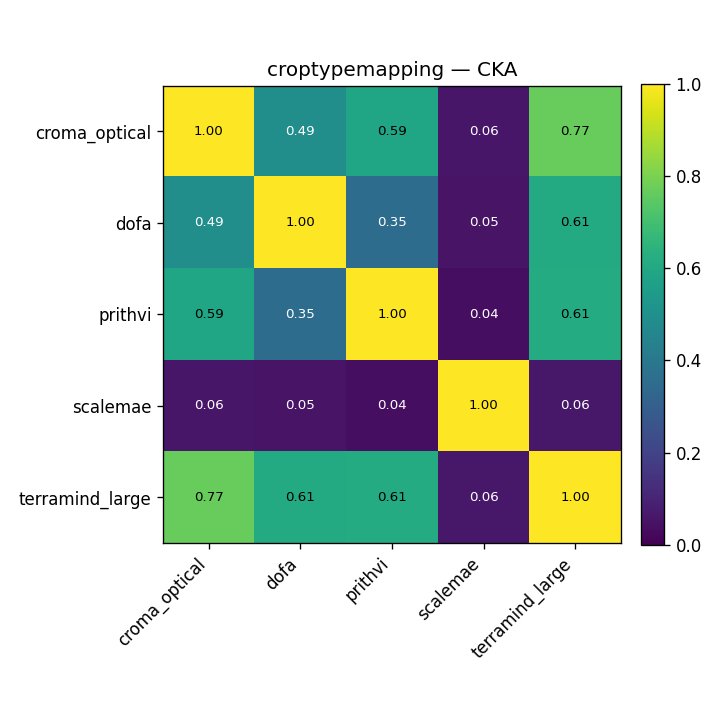}
    \caption{CKA}
    \label{fig:croptypemapping_scores_b}
  \end{subfigure}
  \begin{subfigure}[b]{0.245\linewidth}
    \includegraphics[trim=12 38 11 48, clip, width=\linewidth]{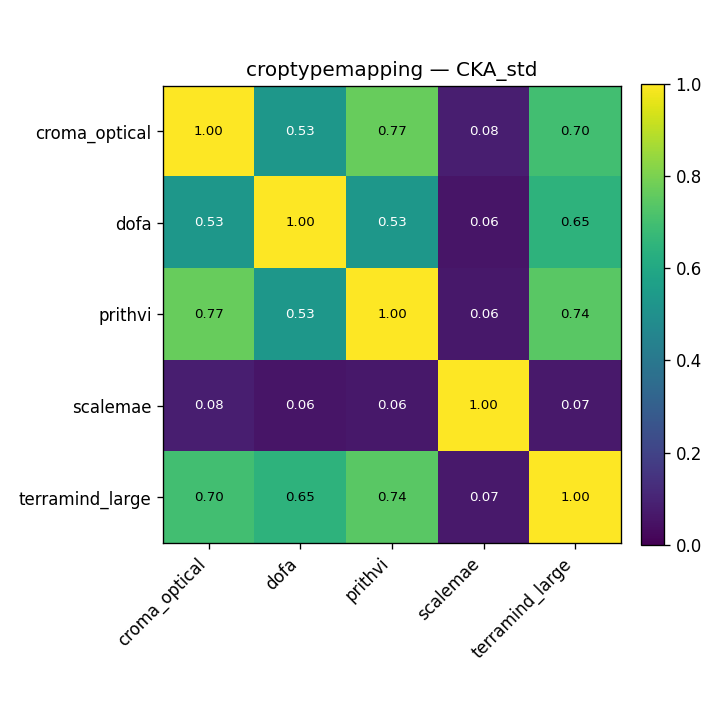}
    \caption{Standardized CKA}
    \label{fig:croptypemapping_scores_c}
  \end{subfigure}
  \begin{subfigure}[b]{0.245\linewidth}
    \includegraphics[trim=12 38 11 52, clip, width=\linewidth]{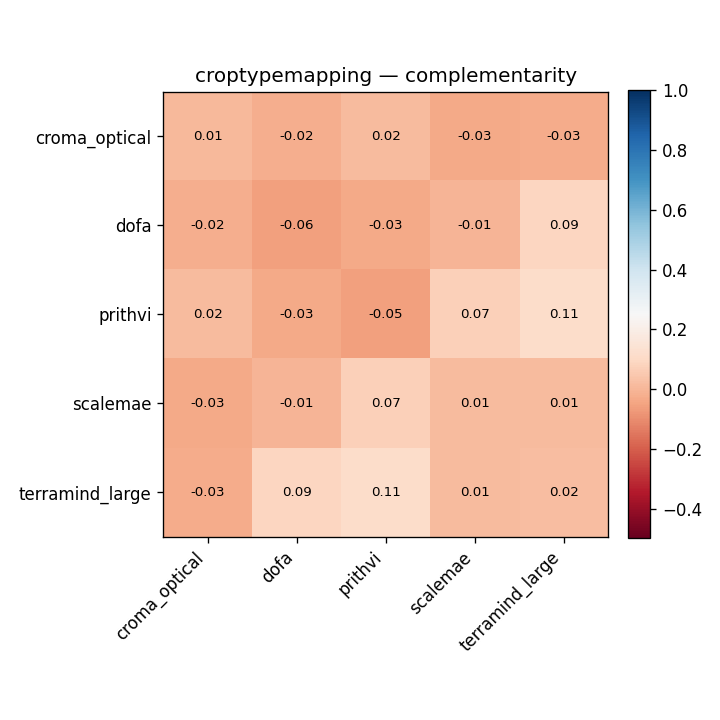}
    \caption{Complementarity}
    \label{fig:croptypemapping_scores_d}
  \end{subfigure}
  \caption{CropTypeMapping similarity and complementarity scores.}
  \label{fig:croptypemapping_scores}
\end{figure}

\begin{figure}[t]
  \centering
  \begin{subfigure}[b]{0.245\linewidth}
    \includegraphics[trim=12 38 11 46, clip, width=\linewidth]{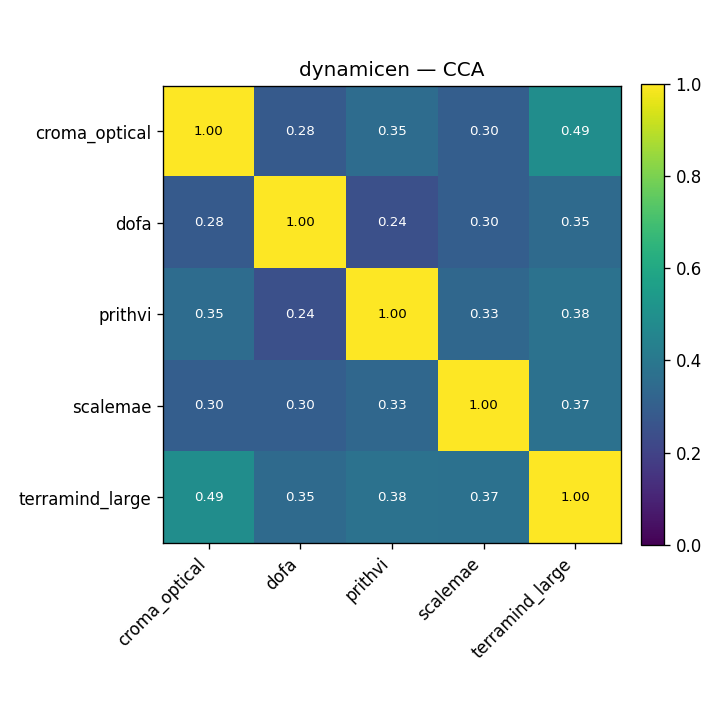}
    \caption{CCA}
    \label{fig:dynamicen_scores_a}
  \end{subfigure}
  \begin{subfigure}[b]{0.245\linewidth}
    \includegraphics[trim=12 38 11 46, clip, width=\linewidth]{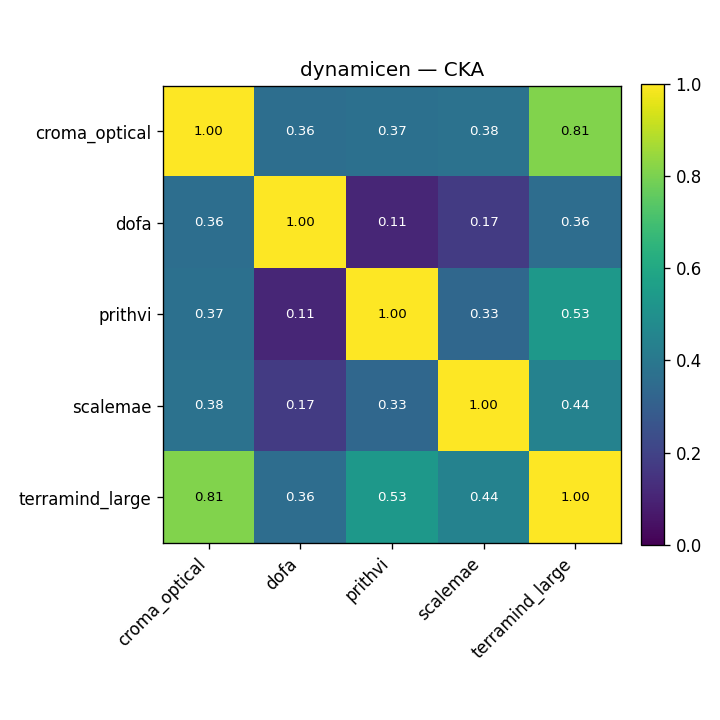}
    \caption{CKA}
    \label{fig:dynamicen_scores_b}
  \end{subfigure}
  \begin{subfigure}[b]{0.245\linewidth}
    \includegraphics[trim=12 38 11 48, clip, width=\linewidth]{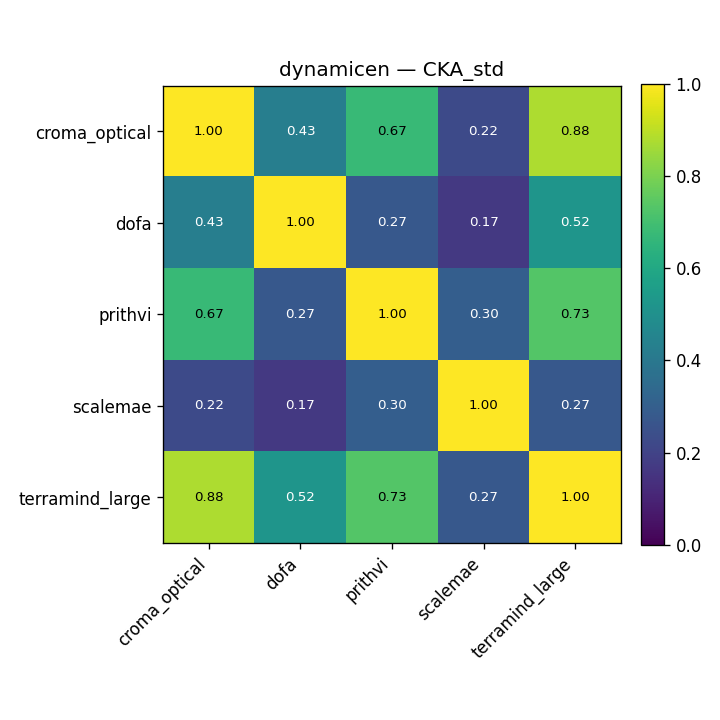}
    \caption{Standardized CKA}
    \label{fig:dynamicen_scores_c}
  \end{subfigure}
  \begin{subfigure}[b]{0.245\linewidth}
    \includegraphics[trim=12 38 11 52, clip, width=\linewidth]{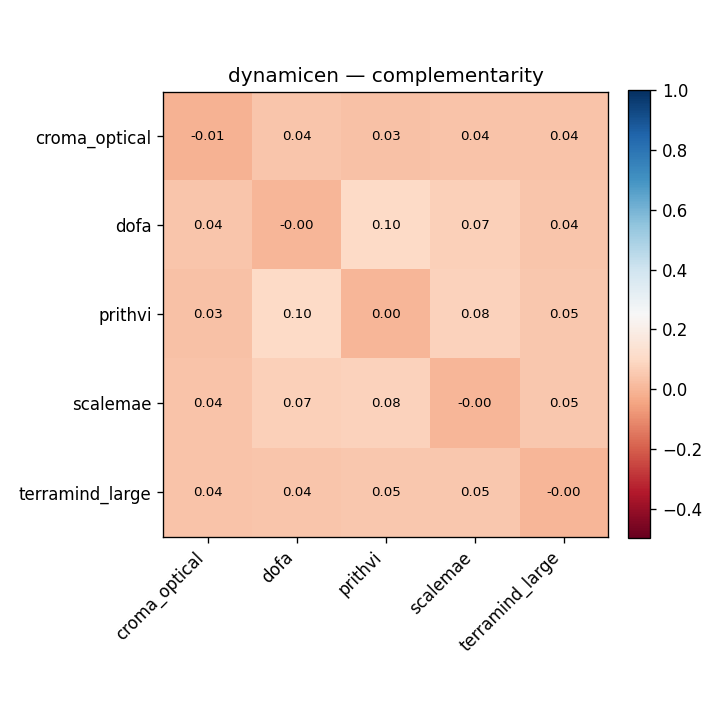}
    \caption{Complementarity}
    \label{fig:dynamicen_scores_d}
  \end{subfigure}
  \caption{DynamicEarthNet similarity and complementarity scores.}
  \label{fig:dynamicen_scores}
\end{figure}

\begin{figure}[t]
  \centering
  \begin{subfigure}[b]{0.245\linewidth}
    \includegraphics[trim=12 38 11 46, clip, width=\linewidth]{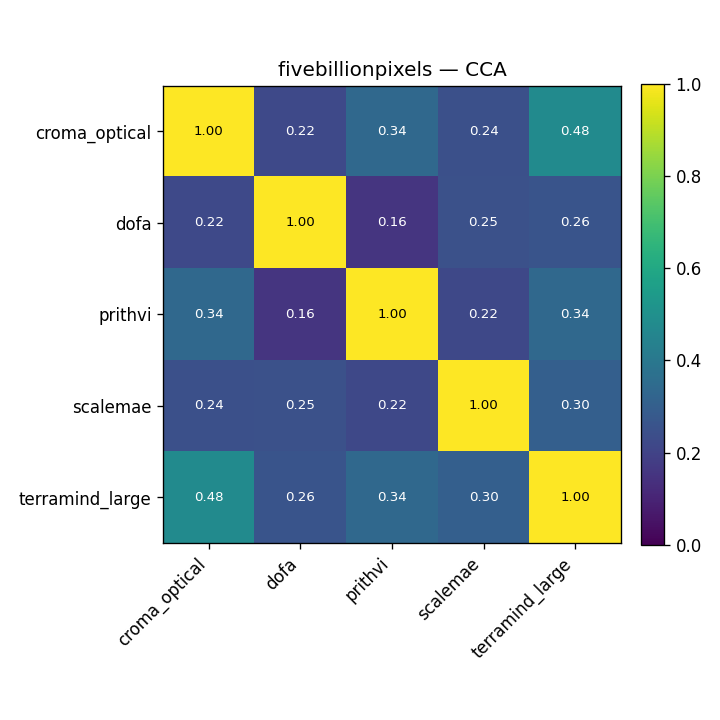}
    \caption{CCA}
    \label{fig:fivebillionpixels_scores_a}
  \end{subfigure}
  \begin{subfigure}[b]{0.245\linewidth}
    \includegraphics[trim=12 38 11 46, clip, width=\linewidth]{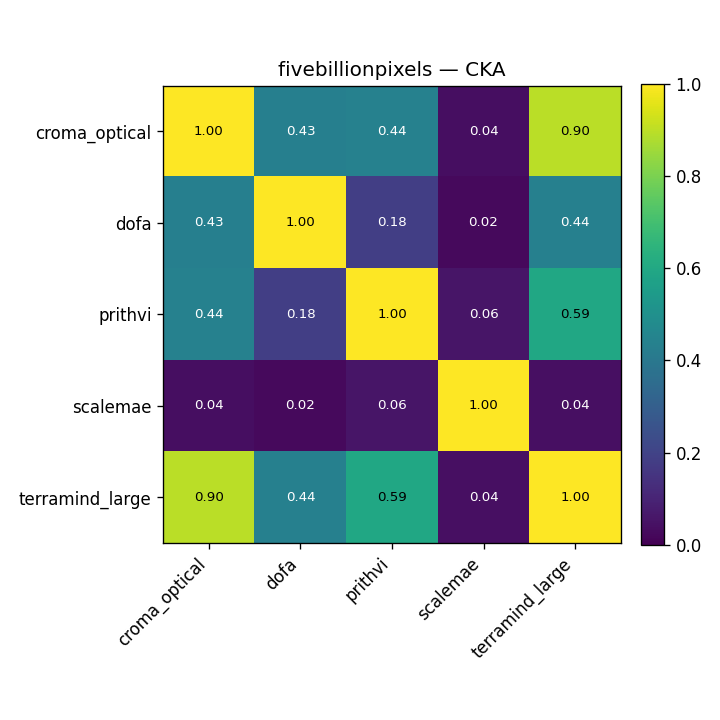}
    \caption{CKA}
    \label{fig:fivebillionpixels_scores_b}
  \end{subfigure}
  \begin{subfigure}[b]{0.245\linewidth}
    \includegraphics[trim=12 38 11 48, clip, width=\linewidth]{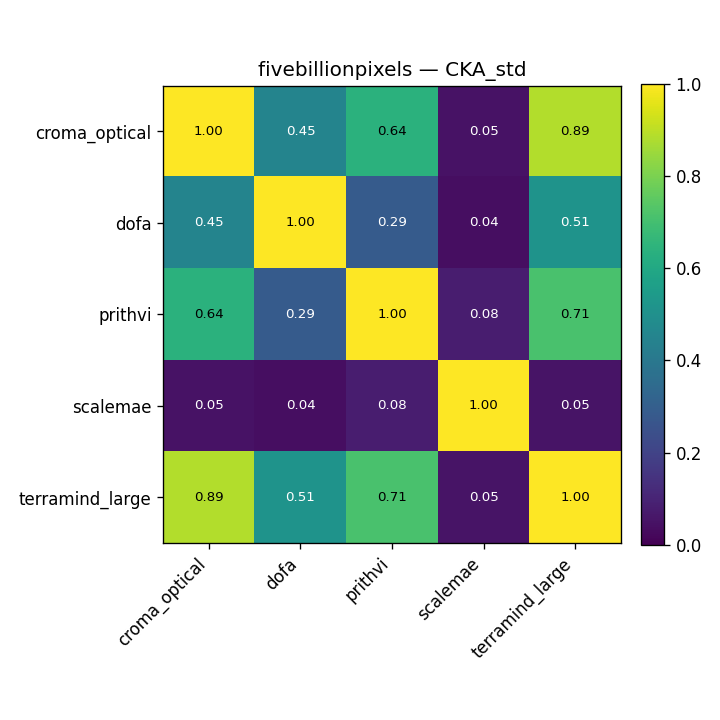}
    \caption{Standardized CKA}
    \label{fig:fivebillionpixels_scores_c}
  \end{subfigure}
  \begin{subfigure}[b]{0.245\linewidth}
    \includegraphics[trim=12 38 11 52, clip, width=\linewidth]{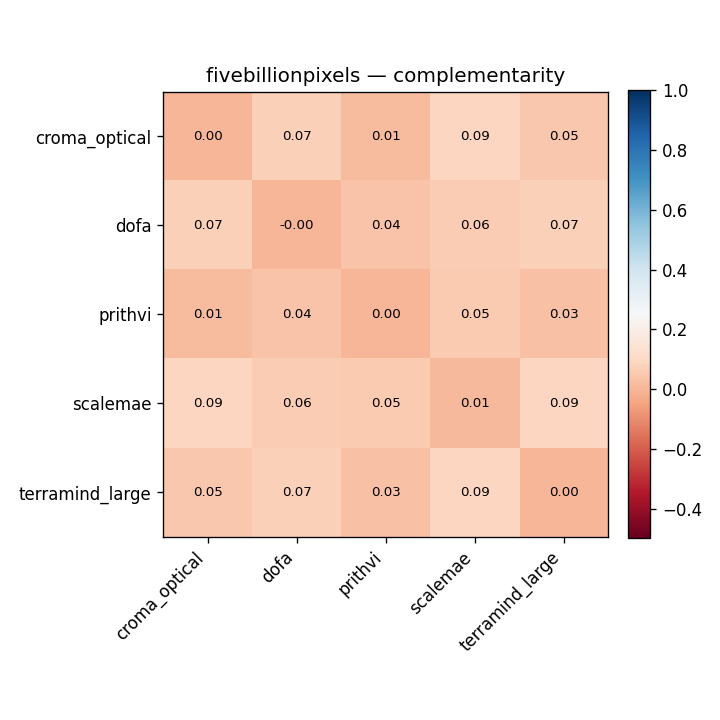}
    \caption{Complementarity}
    \label{fig:fivebillionpixels_scores_d}
  \end{subfigure}
  \caption{FiveBillionPixels similarity and complementarity scores.}
  \label{fig:fivebillionpixels_scores}
\end{figure}

\begin{figure}[t]
  \centering
  \begin{subfigure}[b]{0.245\linewidth}
    \includegraphics[trim=12 38 11 46, clip, width=\linewidth]{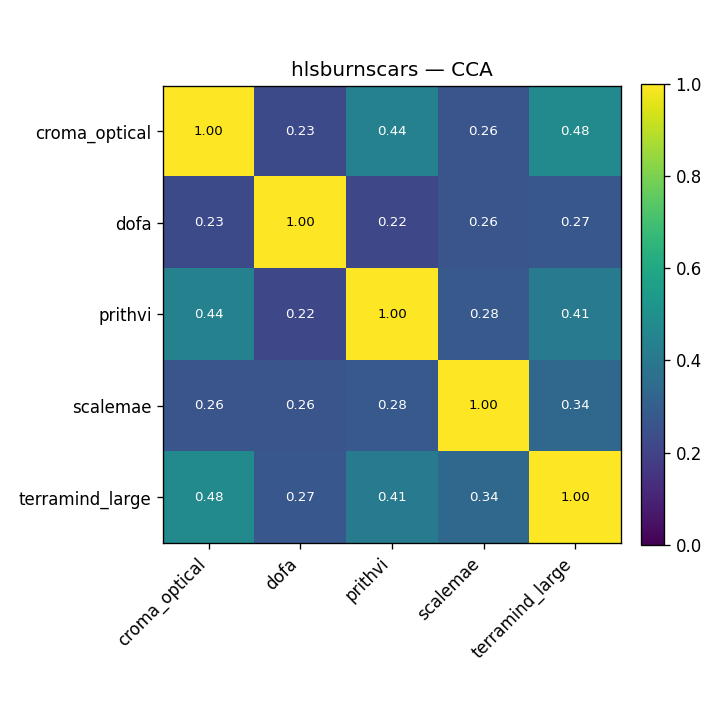}
    \caption{CCA}
    \label{fig:hlsburnscars_scores_a}
  \end{subfigure}
  \begin{subfigure}[b]{0.245\linewidth}
    \includegraphics[trim=12 38 11 46, clip, width=\linewidth]{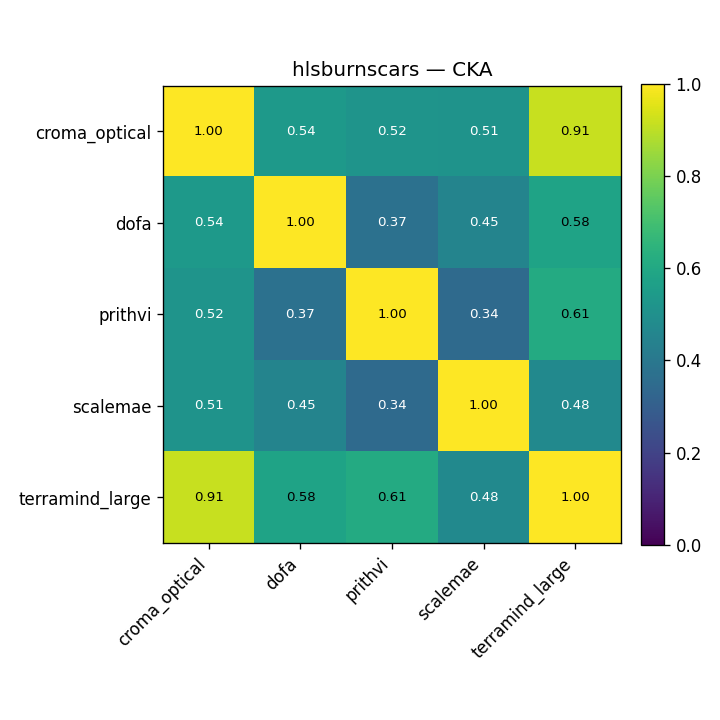}
    \caption{CKA}
    \label{fig:hlsburnscars_scores_b}
  \end{subfigure}
  \begin{subfigure}[b]{0.245\linewidth}
    \includegraphics[trim=12 38 11 48, clip, width=\linewidth]{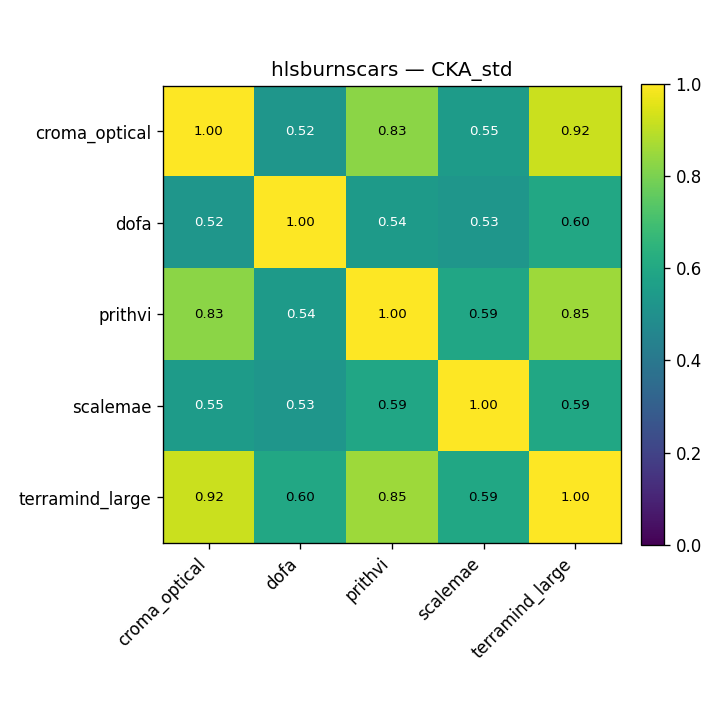}
    \caption{Standardized CKA}
    \label{fig:hlsburnscars_scores_c}
  \end{subfigure}
  \begin{subfigure}[b]{0.245\linewidth}
    \includegraphics[trim=12 38 11 52, clip, width=\linewidth]{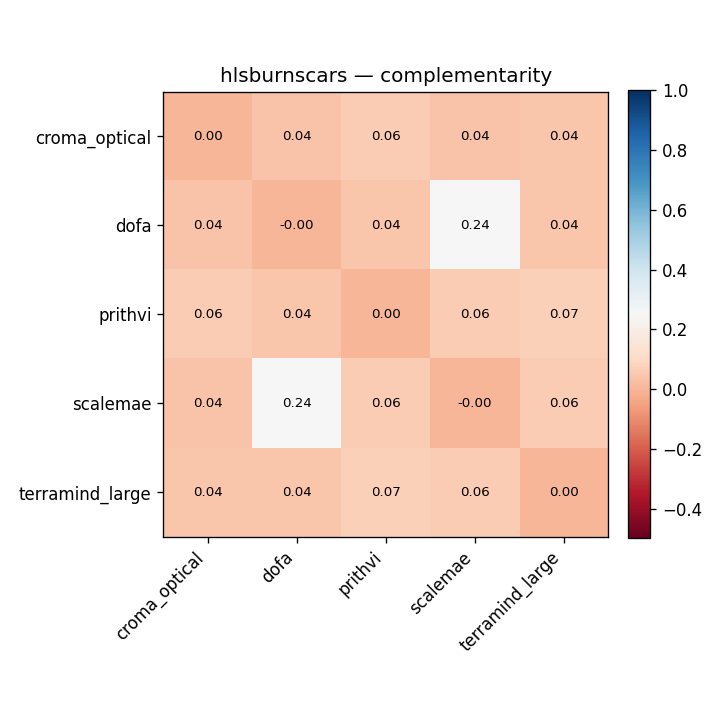}
    \caption{Complementarity}
    \label{fig:hlsburnscars_scores_d}
  \end{subfigure}
  \caption{HLSBurnScars similarity and complementarity scores.}
  \label{fig:hlsburnscars_scores}
\end{figure}

\begin{figure}[t]
  \centering
  \begin{subfigure}[b]{0.245\linewidth}
    \includegraphics[trim=12 38 11 46, clip, width=\linewidth]{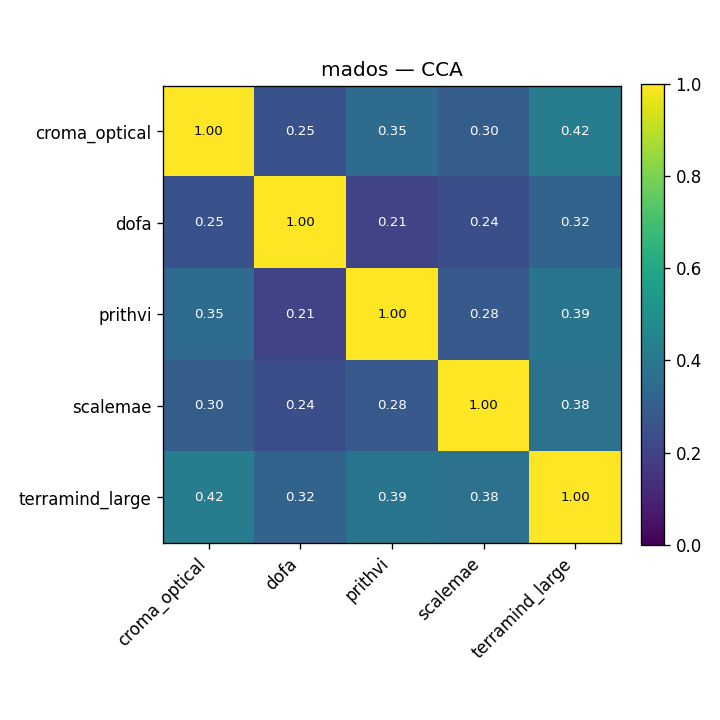}
    \caption{CCA}
    \label{fig:mados_scores_a}
  \end{subfigure}
  \begin{subfigure}[b]{0.245\linewidth}
    \includegraphics[trim=12 38 11 46, clip, width=\linewidth]{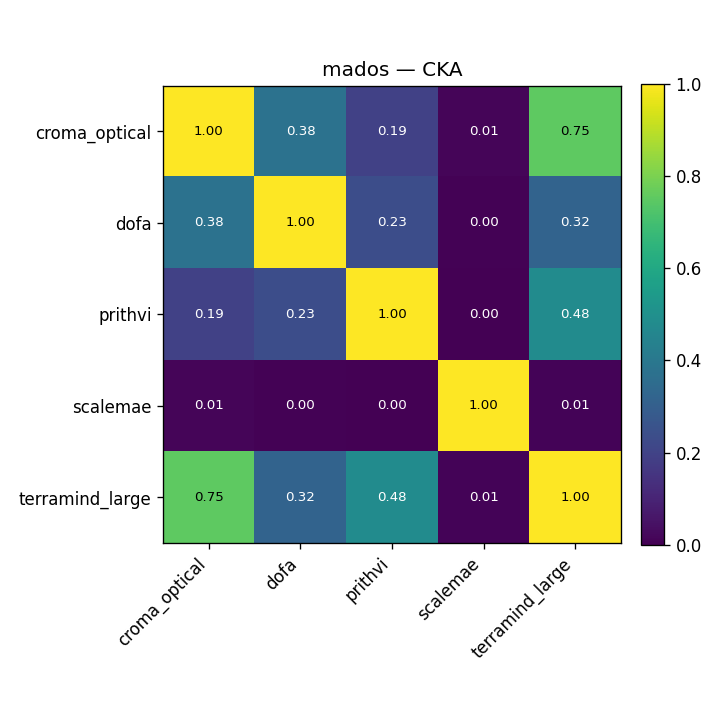}
    \caption{CKA}
    \label{fig:mados_scores_b}
  \end{subfigure}
  \begin{subfigure}[b]{0.245\linewidth}
    \includegraphics[trim=12 38 11 48, clip, width=\linewidth]{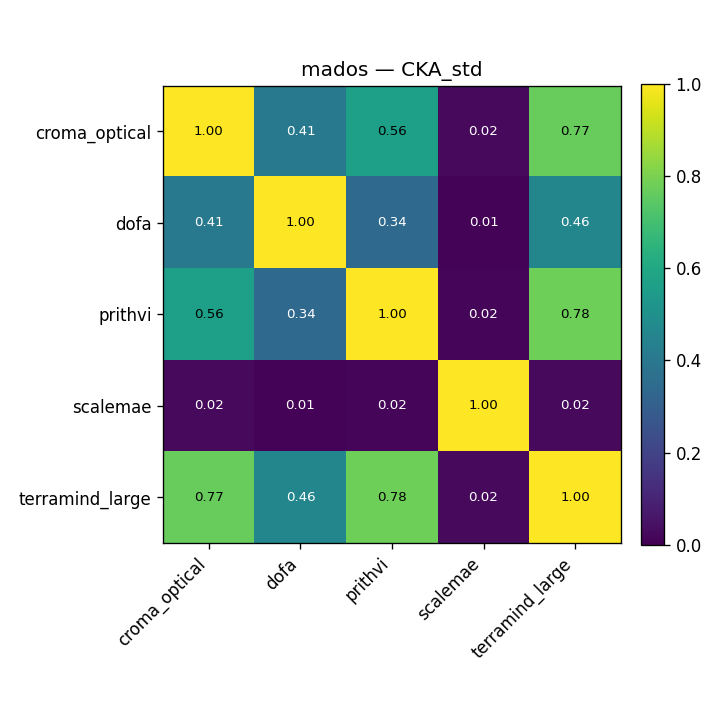}
    \caption{Standardized CKA}
    \label{fig:mados_scores_c}
  \end{subfigure}
  \begin{subfigure}[b]{0.245\linewidth}
    \includegraphics[trim=12 38 11 52, clip, width=\linewidth]{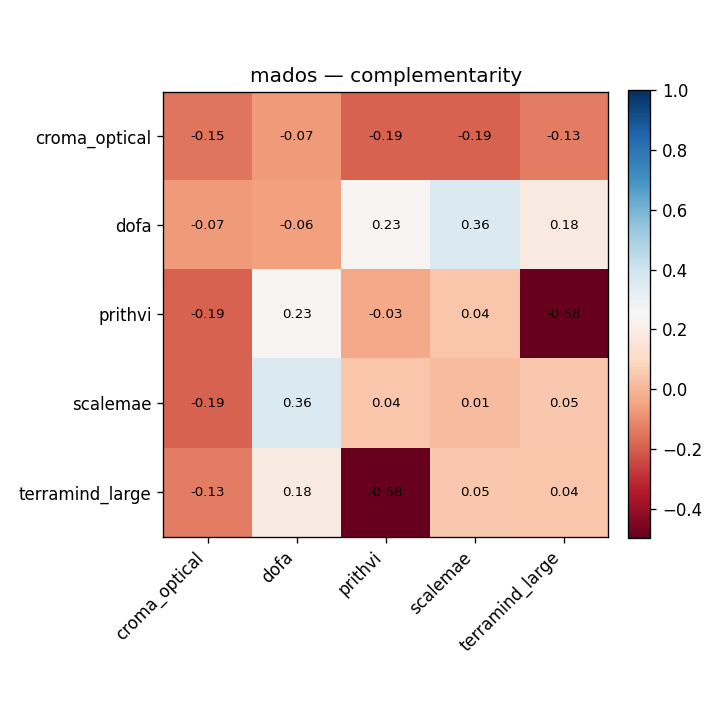}
    \caption{Complementarity}
    \label{fig:mados_scores_d}
  \end{subfigure}
  \caption{MADOS similarity and complementarity scores.}
  \label{fig:mados_scores}
\end{figure}

\begin{figure}[t]
  \centering
  \begin{subfigure}[b]{0.245\linewidth}
    \includegraphics[trim=12 38 11 46, clip, width=\linewidth]{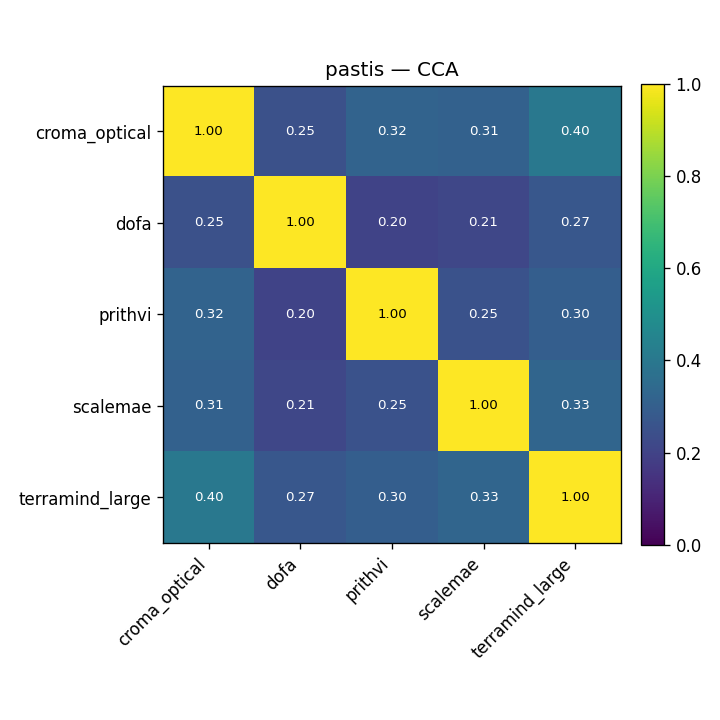}
    \caption{CCA}
    \label{fig:pastis_scores_a}
  \end{subfigure}
  \begin{subfigure}[b]{0.245\linewidth}
    \includegraphics[trim=12 38 11 46, clip, width=\linewidth]{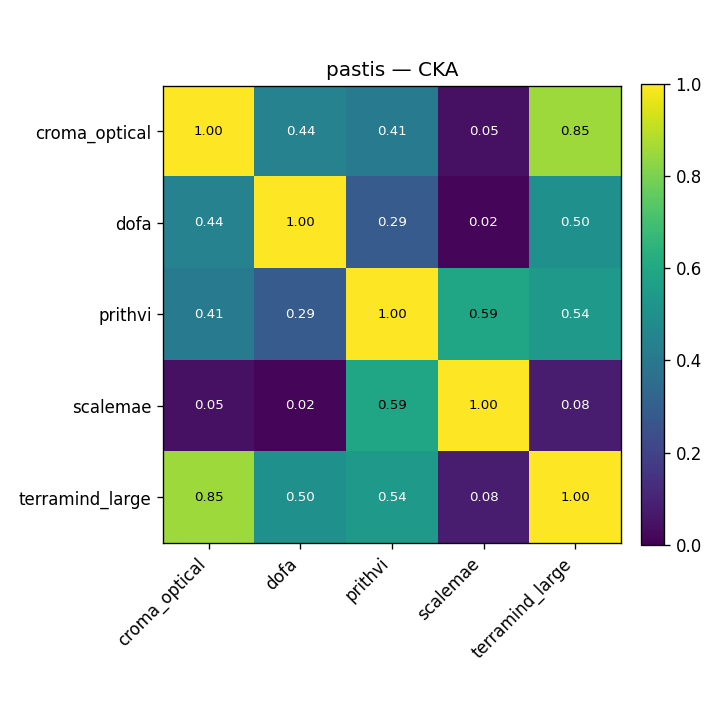}
    \caption{CKA}
    \label{fig:pastis_scores_b}
  \end{subfigure}
  \begin{subfigure}[b]{0.245\linewidth}
    \includegraphics[trim=12 38 11 48, clip, width=\linewidth]{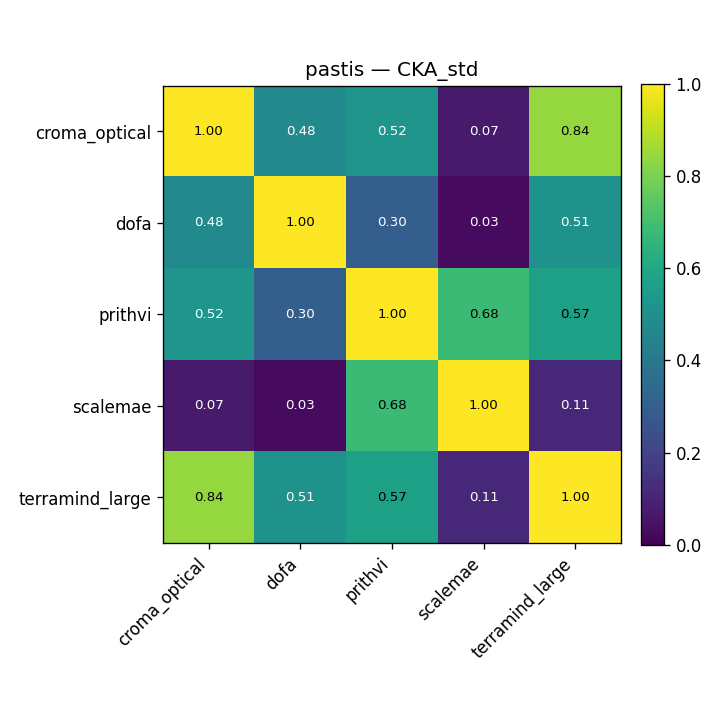}
    \caption{Standardized CKA}
    \label{fig:pastis_scores_c}
  \end{subfigure}
  \begin{subfigure}[b]{0.245\linewidth}
    \includegraphics[trim=12 38 11 52, clip, width=\linewidth]{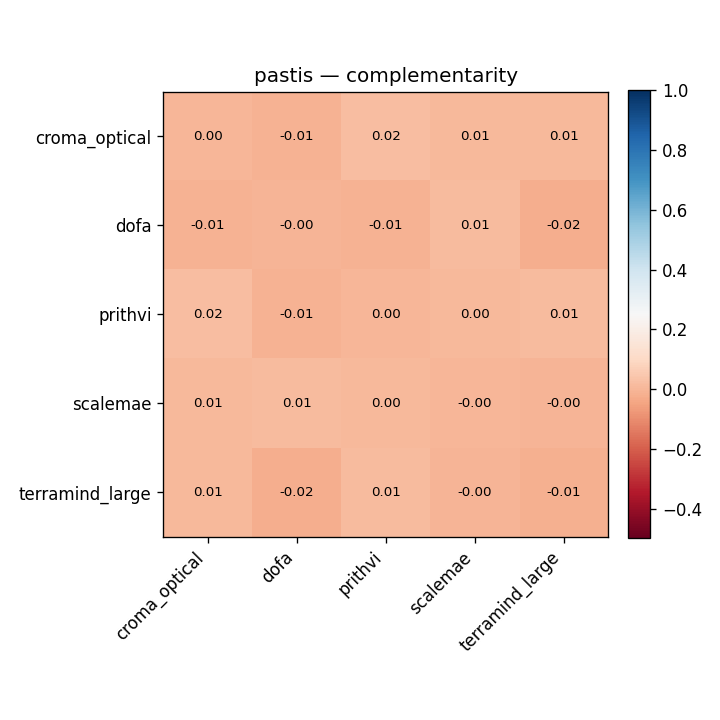}
    \caption{Complementarity}
    \label{fig:pastis_scores_d}
  \end{subfigure}
  \caption{PASTIS similarity and complementarity scores.}
  \label{fig:pastis_scores}
\end{figure}

\begin{figure}[t]
  \centering
  \begin{subfigure}[b]{0.245\linewidth}
    \includegraphics[trim=12 38 11 46, clip, width=\linewidth]{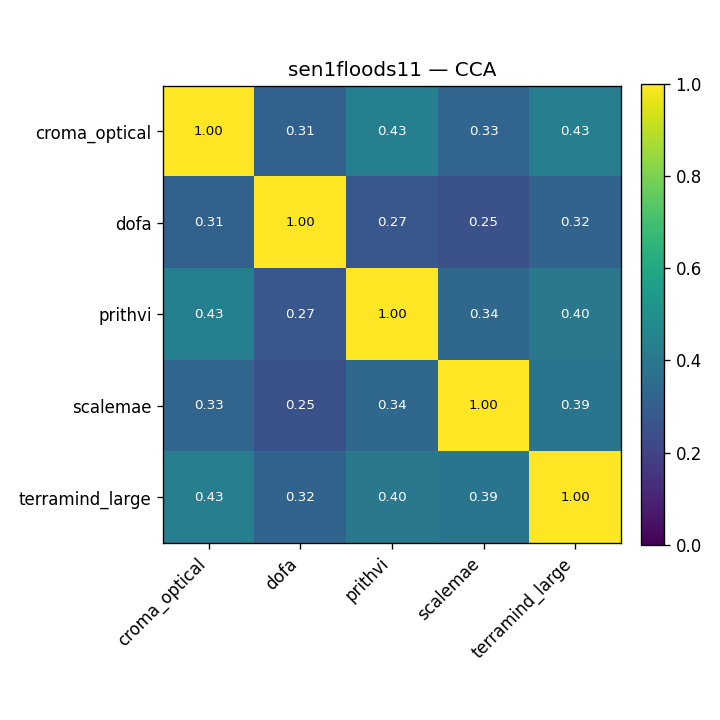}
    \caption{CCA}
    \label{fig:sen1floods11_scores_a}
  \end{subfigure}
  \begin{subfigure}[b]{0.245\linewidth}
    \includegraphics[trim=12 38 11 46, clip, width=\linewidth]{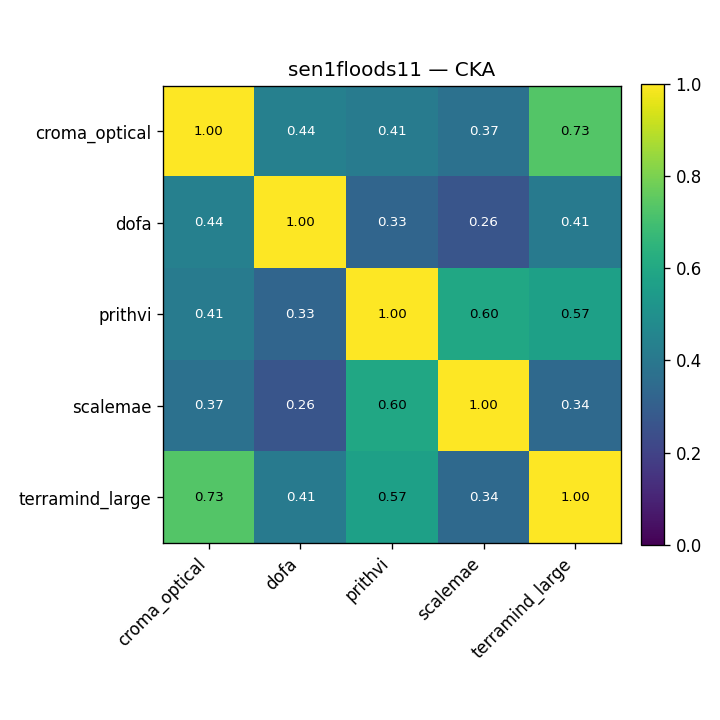}
    \caption{CKA}
    \label{fig:sen1floods11_scores_b}
  \end{subfigure}
  \begin{subfigure}[b]{0.245\linewidth}
    \includegraphics[trim=12 38 11 48, clip, width=\linewidth]{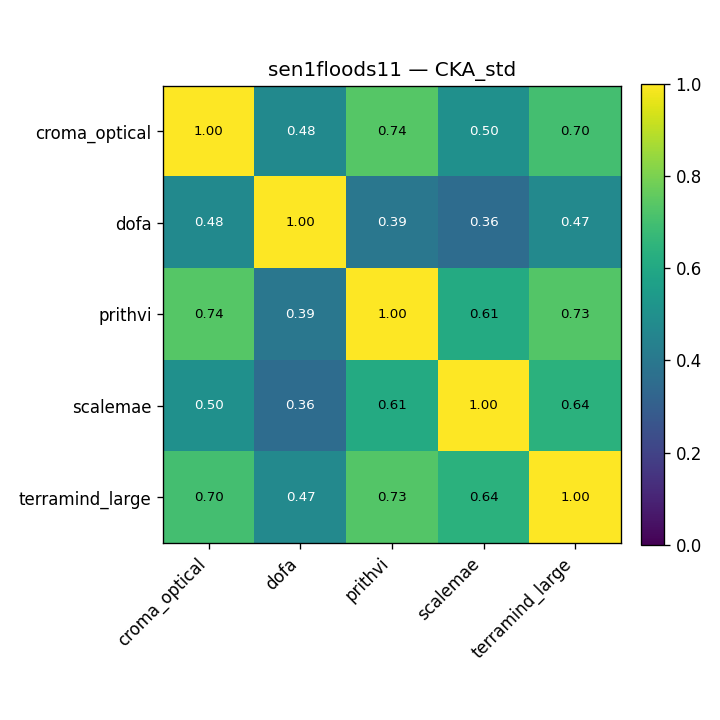}
    \caption{Standardized CKA}
    \label{fig:sen1floods11_scores_c}
  \end{subfigure}
  \begin{subfigure}[b]{0.245\linewidth}
    \includegraphics[trim=12 38 11 52, clip, width=\linewidth]{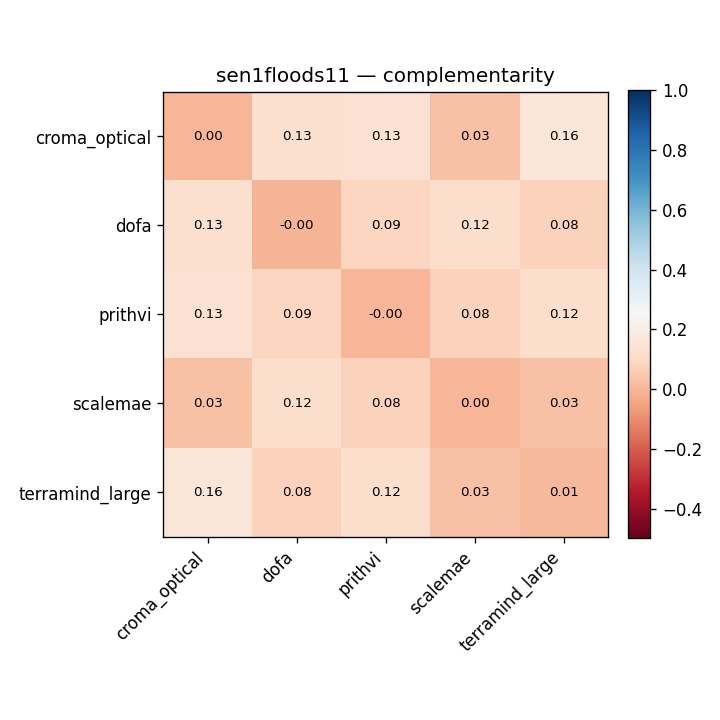}
    \caption{Complementarity}
    \label{fig:sen1floods11_scores_d}
  \end{subfigure}
  \caption{Sen1Floods11 similarity and complementarity scores.}
  \label{fig:sen1floods11_scores}
\end{figure}

\begin{figure}[t]
  \centering
  \begin{subfigure}[b]{0.245\linewidth}
    \includegraphics[trim=12 38 11 46, clip, width=\linewidth]{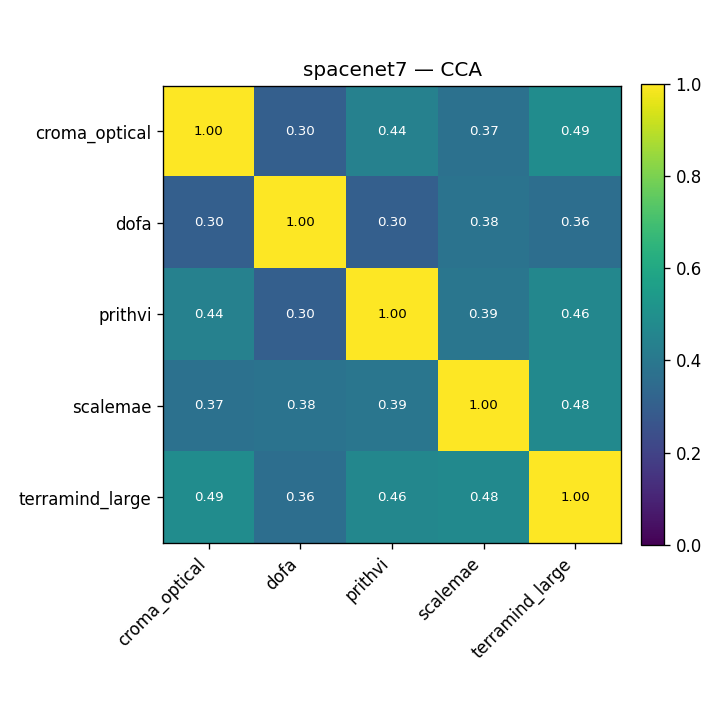}
    \caption{CCA}
    \label{fig:spacenet7_scores_a}
  \end{subfigure}
  \begin{subfigure}[b]{0.245\linewidth}
    \includegraphics[trim=12 38 11 46, clip, width=\linewidth]{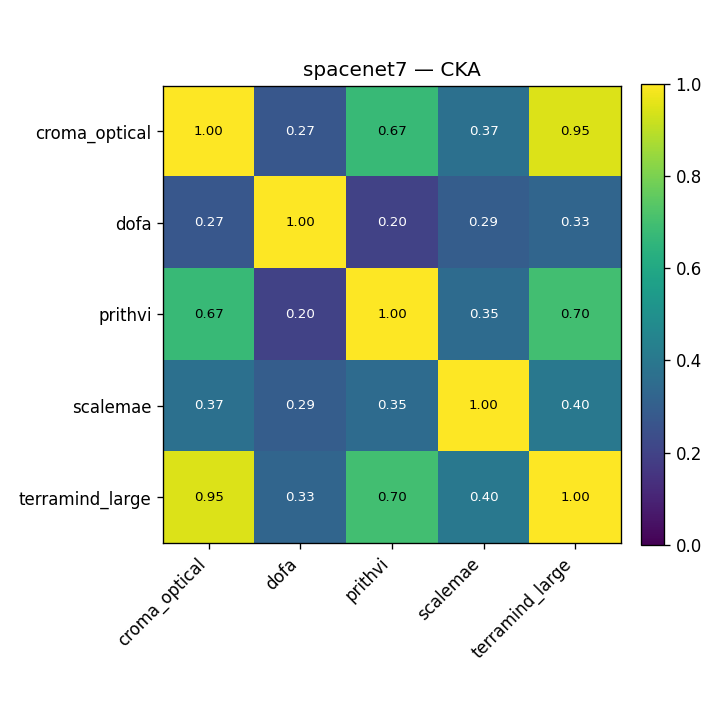}
    \caption{CKA}
    \label{fig:spacenet7_scores_b}
  \end{subfigure}
  \begin{subfigure}[b]{0.245\linewidth}
    \includegraphics[trim=12 38 11 48, clip, width=\linewidth]{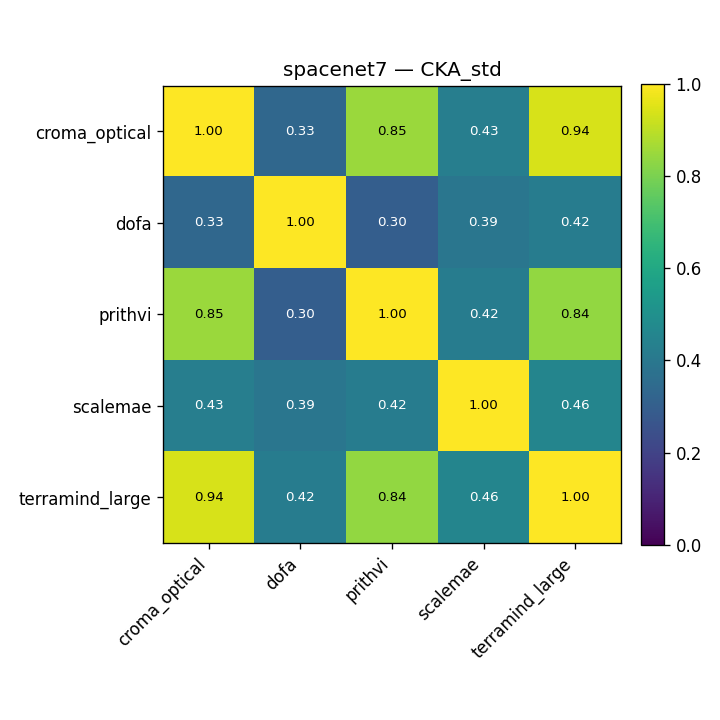}
    \caption{Standardized CKA}
    \label{fig:spacenet7_scores_c}
  \end{subfigure}
  \begin{subfigure}[b]{0.245\linewidth}
    \includegraphics[trim=12 38 11 52, clip, width=\linewidth]{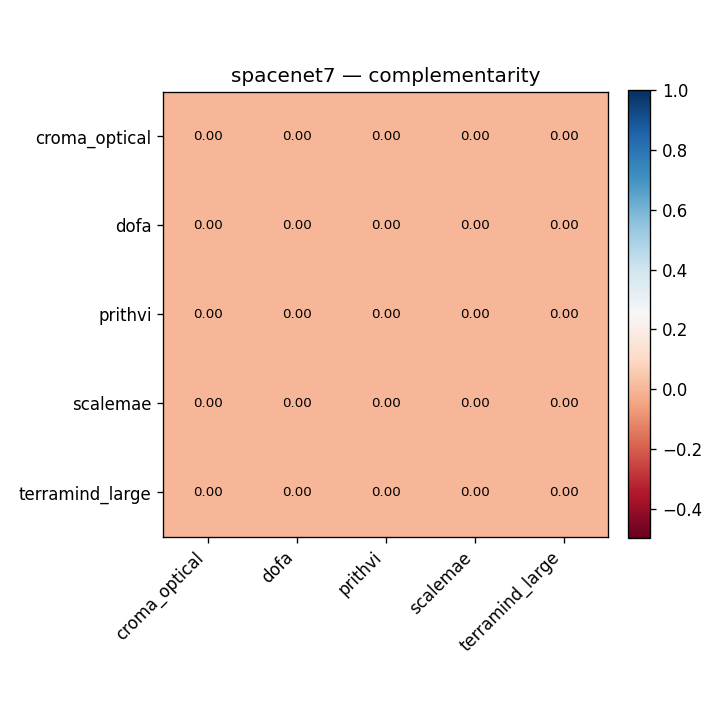}
    \caption{Complementarity}
    \label{fig:spacenet7_scores_d}
  \end{subfigure}
  \caption{SpaceNet7 similarity and complementarity scores.}
  \label{fig:spacenet7_scores}
\end{figure}

\begin{figure}[t]
  \centering
  \begin{subfigure}[b]{0.245\linewidth}
    \includegraphics[trim=12 38 11 46, clip, width=\linewidth]{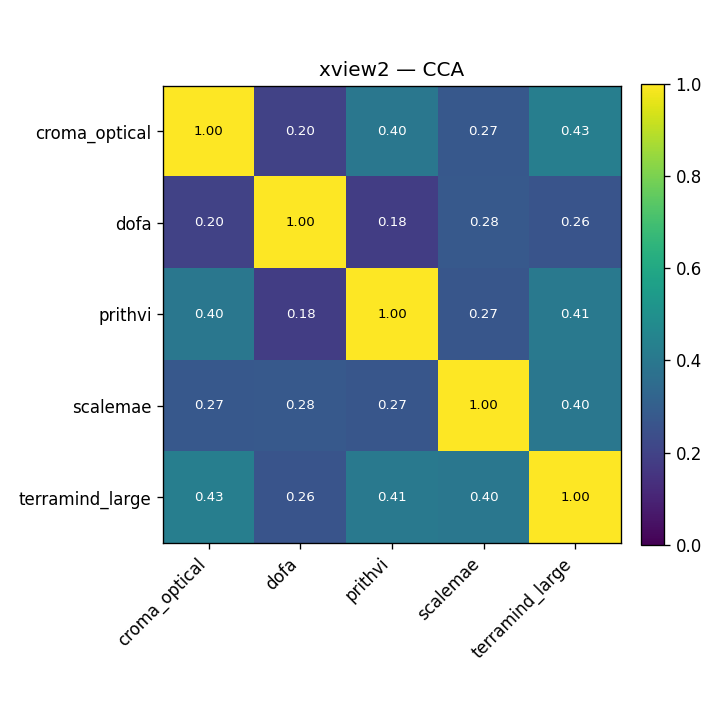}
    \caption{CCA}
    \label{fig:xview2_scores_a}
  \end{subfigure}
  \begin{subfigure}[b]{0.245\linewidth}
    \includegraphics[trim=12 38 11 46, clip, width=\linewidth]{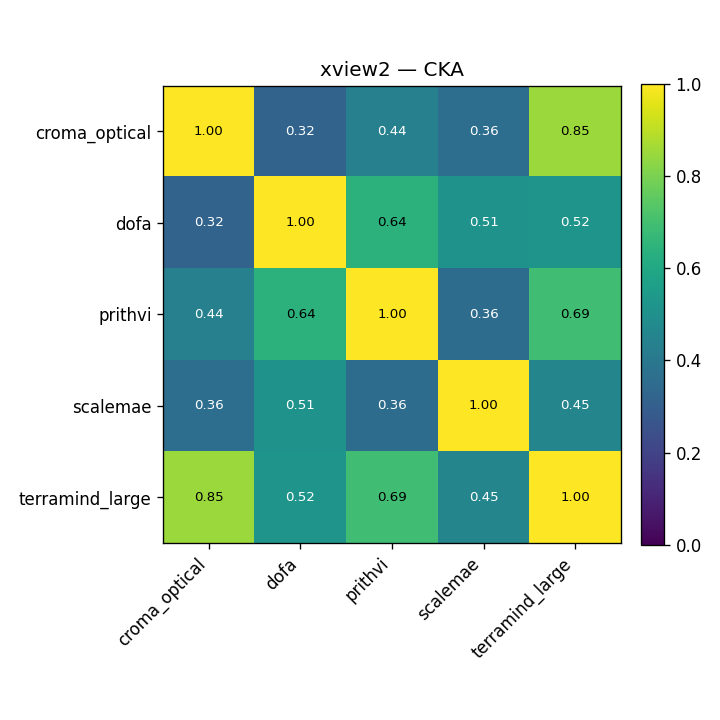}
    \caption{CKA}
    \label{fig:xview2_scores_b}
  \end{subfigure}
  \begin{subfigure}[b]{0.245\linewidth}
    \includegraphics[trim=12 38 11 48, clip, width=\linewidth]{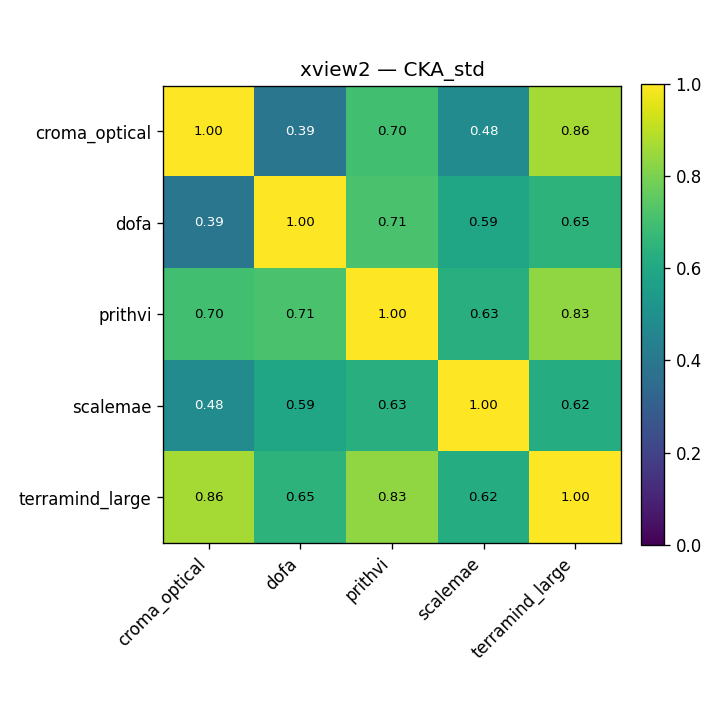}
    \caption{Standardized CKA}
    \label{fig:xview2_scores_c}
  \end{subfigure}
  \begin{subfigure}[b]{0.245\linewidth}
    \includegraphics[trim=12 38 11 52, clip, width=\linewidth]{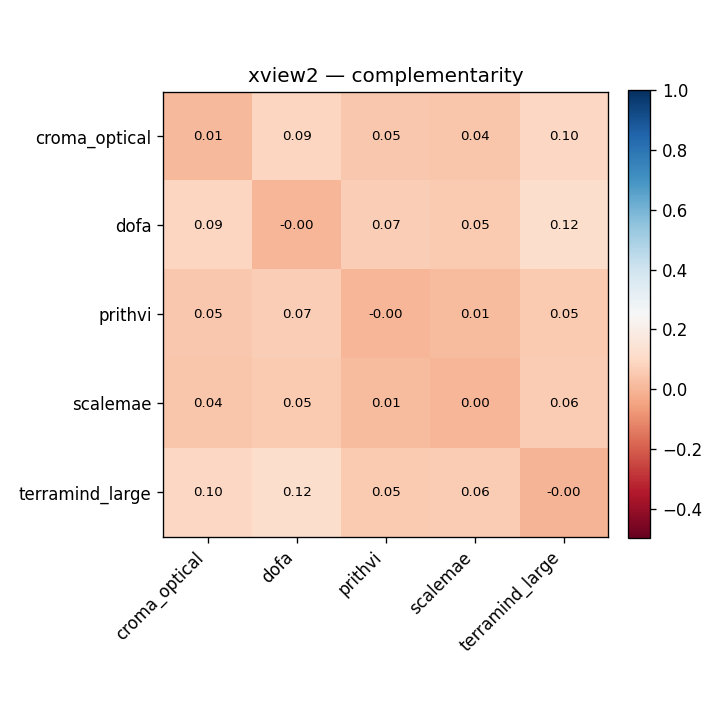}
    \caption{Complementarity}
    \label{fig:xview2_scores_d}
  \end{subfigure}
  \caption{xView2 similarity and complementarity scores.}
  \label{fig:xview2_scores}
\end{figure}

\begin{figure}[t]
  \centering
  \begin{subfigure}[b]{0.245\linewidth}
    \includegraphics[trim=12 38 11 46, clip, width=\linewidth]{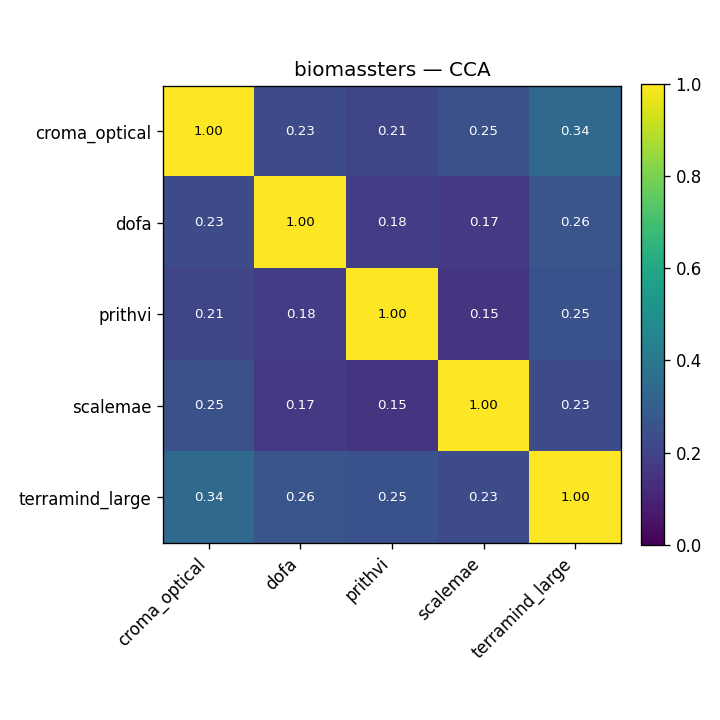}
    \caption{CCA}
    \label{fig:biomassters_scores_a}
  \end{subfigure}
  \begin{subfigure}[b]{0.245\linewidth}
    \includegraphics[trim=12 38 11 46, clip, width=\linewidth]{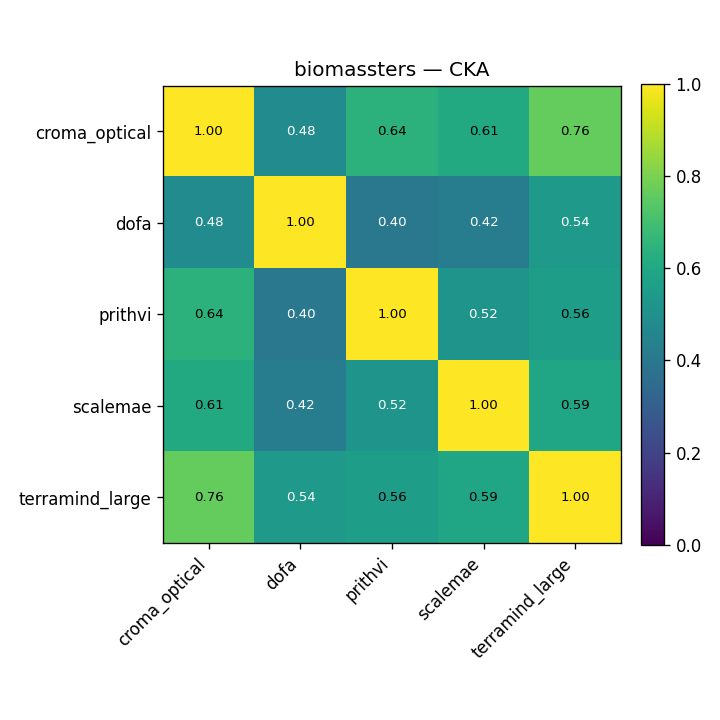}
    \caption{CKA}
    \label{fig:biomassters_scores_b}
  \end{subfigure}
  \begin{subfigure}[b]{0.245\linewidth}
    \includegraphics[trim=12 38 11 48, clip, width=\linewidth]{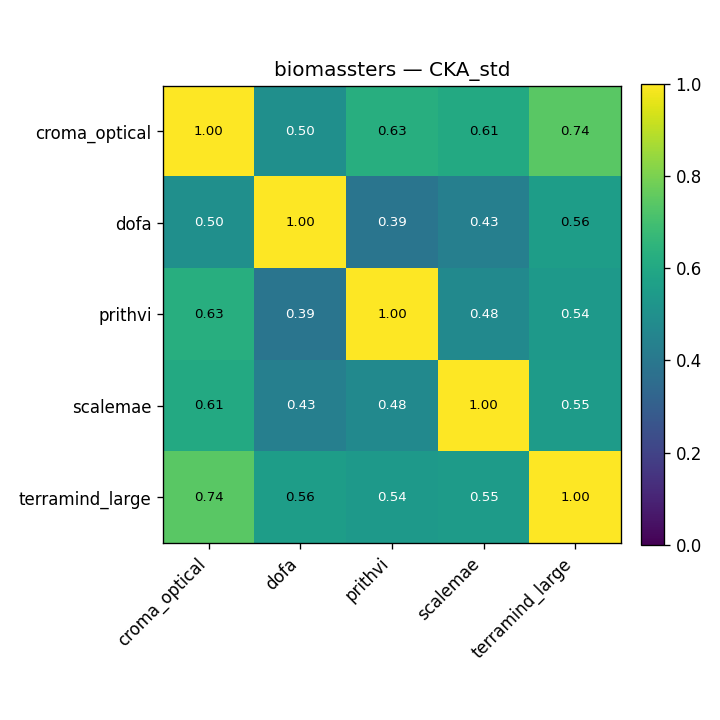}
    \caption{Standardized CKA}
    \label{fig:biomassters_scores_c}
  \end{subfigure}
  \begin{subfigure}[b]{0.245\linewidth}
    \includegraphics[trim=12 38 11 52, clip, width=\linewidth]{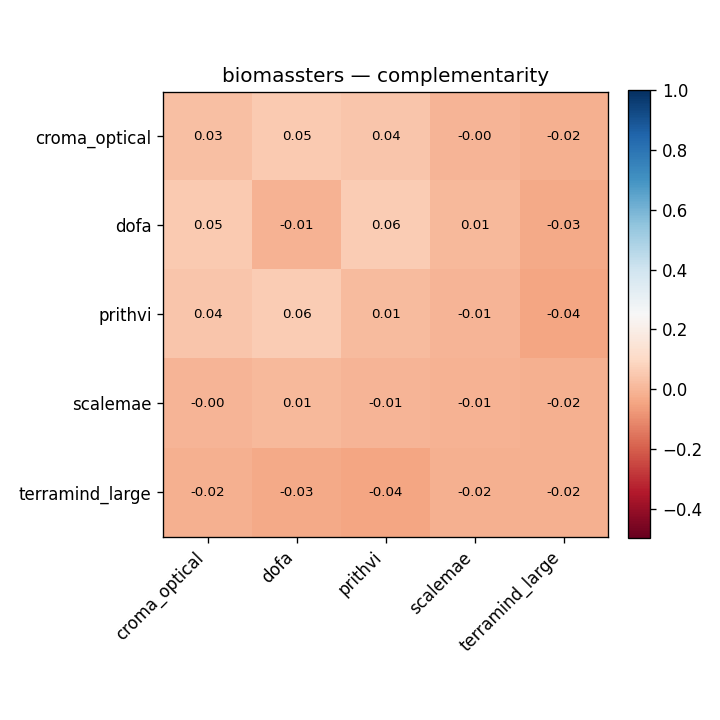}
    \caption{Complementarity}
    \label{fig:biomassters_scores_d}
  \end{subfigure}
  \caption{BioMassters similarity and complementarity scores.}
  \label{fig:biomassters_scores}
\end{figure}

\Cref{fig:ai4smallfarms_scores,fig:biomassters_scores,fig:croptypemapping_scores,fig:dynamicen_scores,fig:fivebillionpixels_scores,fig:hlsburnscars_scores,fig:mados_scores,fig:pastis_scores,fig:sen1floods11_scores,fig:spacenet7_scores,fig:xview2_scores} provide quantitative results of the CCA, CKA, standardized CKA and complementarity scores (Sec.~\ref{sec:similarity_complementarity}, App.~\ref{sec:app_similarity_complementarity}).

\end{document}